%% file: main.tex
\documentclass[10pt,twocolumn,letterpaper]{article}

\usepackage[pagenumbers]{iccv} 


\definecolor{iccvblue}{rgb}{0.21,0.49,0.74}

\usepackage[pagebackref,breaklinks,colorlinks,allcolors=iccvblue]{hyperref}
\usepackage{multirow}
\usepackage{booktabs}  
\usepackage{graphicx}  
\usepackage{colortbl}  
\usepackage{algorithm}
\usepackage{algpseudocode}
\usepackage{arydshln}
\def\paperID{1115} 
\def\confName{ICCV}
\def\confYear{2025}

\title{APT: Improving Diffusion Models for High Resolution \\ Image Generation with Adaptive Path Tracing}

\author{
Sangmin Han, Jinho Jeong, Jinwoo Kim, Seon Joo Kim\\
Yonsei University\\
{\tt\small \{smhan213, 3587jjh, jinwoo-kim, seonjookim\}@yonsei.ac.kr}\\
}

\begin{document}

\twocolumn[{%
\renewcommand\twocolumn[1][]{#1}%
\maketitle
\input{partial/fig_teaser}
}]

\input{section/0_abstract}    
\input{section/1_introduction}
\input{section/2_related_work}
\input{section/3_observations}
\input{section/4_method}
\input{section/5_experiments}

\input{section/6_conclusion}

\input{supple}
\clearpage
{
    \small
    \bibliographystyle{ieeenat_fullname}
    \bibliography{main}
}


\end{document}

%% file: partial/fig_teaser.tex
\vspace{-0.8cm} 
\begin{center}
    \centering
    \includegraphics[width=\textwidth]{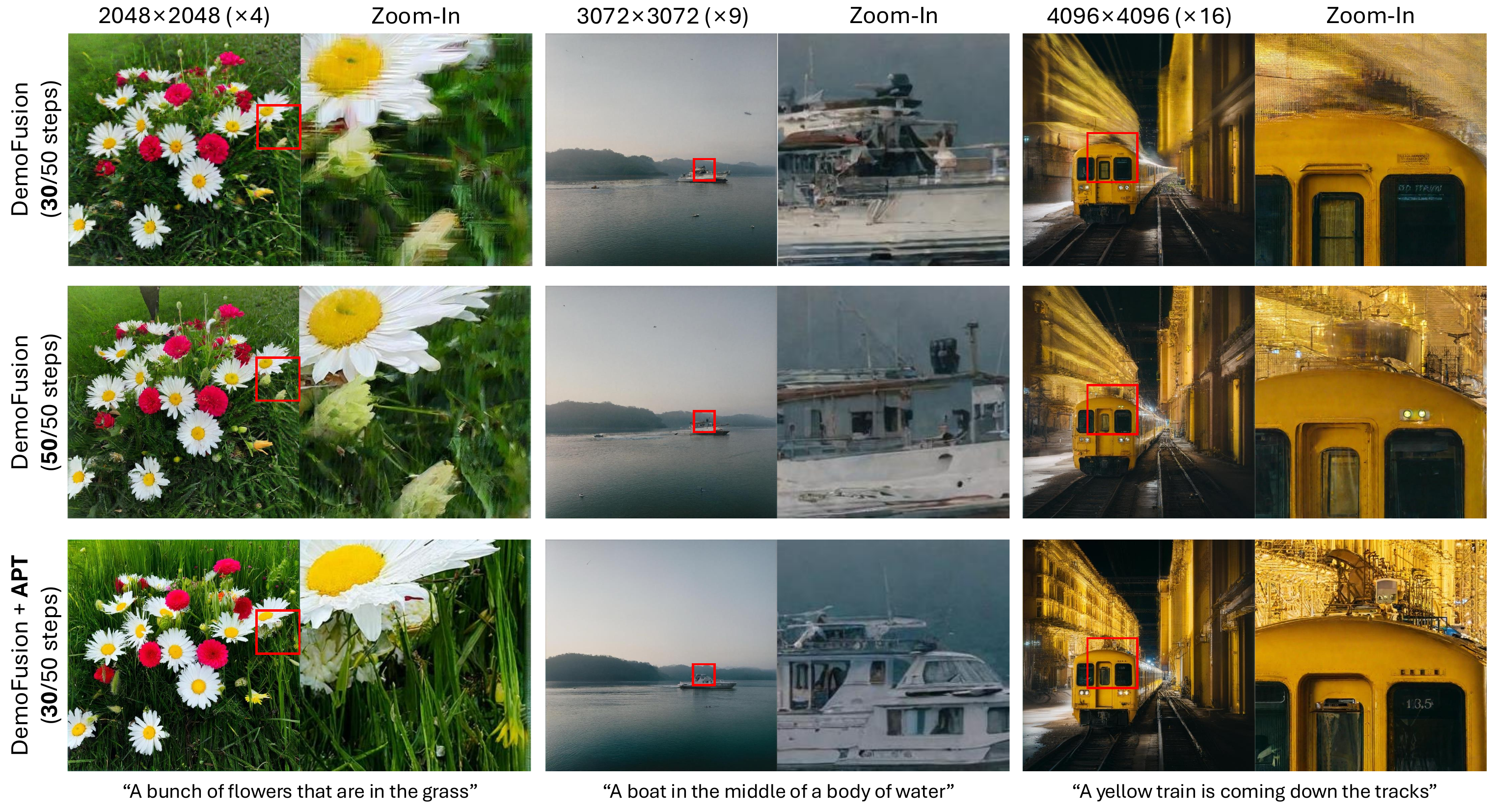}  
    \captionof{figure}{
        \textbf{Comparison of high-resolution image generation using DemoFusion with and without APT.}
        Our APT (Adaptive Path Tracing) achieves superior clarity and detail with reduced sampling steps (30/50), demonstrating both efficiency and effectiveness in high-resolution image generation.}
    \label{fig:teaser}
\end{center}

%% file: section/0_abstract.tex
\begin{abstract}

Latent Diffusion Models (LDMs) are generally trained at fixed resolutions, limiting their capability when scaling up to high-resolution images. 
While training-based approaches address this limitation by training on high-resolution datasets, they require large amounts of data and considerable computational resources, making them less practical.
Consequently, training-free methods, particularly patch-based approaches, have become a popular alternative. 
These methods divide an image into patches and fuse the denoising paths of each patch, 
showing strong performance on high-resolution generation. 
However, we observe two critical issues for patch-based approaches, which we call ``patch-level distribution shift" and ``increased patch monotonicity." 
To address these issues, we propose Adaptive Path Tracing (APT), a framework that combines Statistical Matching to ensure patch distributions remain consistent in upsampled latents and Scale-aware Scheduling to deal with the patch monotonicity.
As a result, APT produces clearer and more refined details in high-resolution images.
In addition, APT enables a shortcut denoising process, resulting in faster sampling with minimal quality degradation.
Our experimental results confirm that APT produces more detailed outputs with improved inference speed, providing a practical approach to high-resolution image generation.
\end{abstract}


%% file: section/1_introduction.tex
\section{Introduction}
\label{sec:intro}

Diffusion models have introduced a new paradigm in generative tasks, demonstrating exceptional capabilities in image generation~\cite{meng2021sdedit, rombach2022high, zhang2023adding, hertz2022prompt, brooks2023instructpix2pix, zhao2024uni}. 
Despite their success, the substantial computational cost for training on high-resolution image datasets poses a challenge.
Latent Diffusion Models (LDMs) utilize a low-resolution latent space to reduce computational demands~\cite{rombach2022high, podellsdxl}, enabling the generation of images with resolutions of up to $1024^2$. 

To increase the resolution much higher, integrating super-resolution models can be suggested as a solution~\cite{zhang2021designing, kim2024arbitrary}.
Although simple, this approach lacks the ability to generate realistic fine details necessary for high-resolution image generation~\cite{he2023scalecrafter, du2024demofusion}.
Alternatively, fully training diffusion models on target high-resolution image datasets is possible~\cite{crowson2024scalable, ho2022cascaded}. However, such methods require considerable computational resources, which are not readily available.


Recent studies~\cite{jin2024training, he2023scalecrafter, huang2024fouriscale} address the problem in a training-free manner. Among them, the main stream is \textit{patch-based} method for the strong performance~\cite{bar2023multidiffusion, lee2023syncdiffusion, du2024demofusion, lin2025accdiffusion}. 
These methods generate high-resolution images by fusing denoising paths of multiple overlapping patches of the pre-trained size. 
Though multiple patch denoising is time-consuming, each patch contributes intricate local details, which are crucial for high-resolution image generation.

Du et al.~\cite{du2024demofusion} proposed DemoFusion, the foundational patch-based method leveraging “upsample-diffusion-denoising” pipeline. 
This approach has been widely adopted in follow-up studies~\cite{lin2025accdiffusion, shi2024resmaster, tragakis2024one}. 
The pipeline begins by upsampling a latent at a pre-trained resolution using conventional non-parametric interpolation techniques such as bicubic. 
It then refines the upsampled latent through a diffusion-denoising process that fuses local and dilated patches. 
Local patches contain adjacent pixels to generate fine details, whereas dilated patches sample pixels at a fixed stride to enhance global coherence.
These patches have same resolution with the initial latent at the pre-trained resolution. 
For example, when an initial latent of size $64^2$ is upsampled to $128^2$, both local patches and dilated patches are sampled at $64^2$ during denoising process.

Despite its strong performance, we identify two critical issues in the patch-based methods induced by the conventional upsampling: 
\textit{patch-level distribution shift} and \textit{increased 
patch monotonicity} of the upsampled latent.
Ideally, dilated patches should be refined in a way that preserves both their alignment with the initial low-resolution latent and their mutual consistency for global coherence. However, conventional upsampling introduces distribution shifts that undermine both requirements, resulting in inconsistent reconstructions and ultimately degrading the final output.
Meanwhile, increased patch monotonicity arises from increased pixel similarity within a local patch due to its smaller receptive field. 
This increased similarity reduces the signal-to-noise ratio (SNR), preventing local patches from being adaptively diffused and denoised, negatively impacting the final output.
We further validate our insights through toy examples and highlight the need to address these issues in \Cref{sec:observations}.

Based on our observations, we introduce \textbf{APT} (\textbf{A}daptive \textbf{P}ath \textbf{T}racing), which employs two simple and effective techniques.
Firstly, for the pixel-level distribution shift problem, we propose \textit{Statistical Matching}. It is based on our observation that statistics of latents (e.g., mean and variance) play a key role in refining distribution-shifted upsampled latent.
Specifically, we apply statistical matching for dilated patches to adjust the mean and variance of the upscaled latent, aligning its statistical properties with those of the original low-resolution latent. 

Secondly, to tackle the patch monotonicity problem, we introduce \textit{Scale-aware Scheduling} during local patch sampling. 
Our approach is inspired by Hoogeboom et al. (Simple Diffusion)~\cite{hoogeboom2023simple}.
Simple Diffusion relies on the insight that pixel redundancy increases with image resolution, proposing a resolution-aware beta scheduling that adapts to the overall image size.
However, the resolution of patches remains fixed and does not change according to the target high-resolution image in the patch-based framework.
As a result, Simple Diffusion cannot adequately address the variations in pixel redundancy that arise within the fixed-size patches.
Thus, we present a new scheduling strategy to address pixel redundancy within fixed-resolution patches.

Our proposed techniques enable APT to produce highly detailed and realistic high-resolution images while significantly reducing computational costs, achieving a runtime improvement of approximately 40\%. 
We provide extensive quantitative and qualitative evaluations, using appropriate metrics and visualizations to thoroughly assess the fine details in the generated high-resolution images. 
In addition, we perform comprehensive ablation studies to provide deeper insight into the effect of each APT component.

%% file: section/2_related_work.tex
\section{Related work}
\label{sec:related_works}




High-resolution image generation has long been a key challenge in generative modeling. The most straightforward solution is \textit{to train models} with high-resolution images. Recent methods, including SDXL~\cite{podellsdxl} and Matryoshka~\cite{gu2023matryoshka}, leverage efficient architectures to generate images up to $1024^2$ resolution, while Pixart-$\Sigma$~\cite{chen2024pixart} introduces a novel training strategy for 4K image datasets. SelfCascade~\cite{guo2025make} enhances resolution by integrating adapters to each UNet's layer for efficient fine-tuning. However, training-based methods often struggle with data scarcity, high GPU costs, and limited scalability beyond the training resolution.

\input{partial/fig_mean_variance}

To address these limitations, \textit{training-free methods} have gained attention by utilizing foundation models~\cite{rombach2022high, podellsdxl} to support higher resolutions. These approaches can be categorized into two groups. The first group modifies pre-trained model architectures to directly process high-resolution noise inputs~\cite{he2023scalecrafter, huang2024fouriscale, zhang2023hidiffusion}. While they yield natural results, it is restricted to small upscaling factors (e.g., $\times$4).

The second group is patch-based methods, which generate high-resolution images by fusing patches of the pre-trained resolution. 
MultiDiffusion~\cite{bar2023multidiffusion} and SyncDiffusion~\cite{lee2023syncdiffusion} are the primitive approaches of the patch-based methods.
DemoFusion~\cite{du2024demofusion} further improves the quality by leveraging the “upsample-diffuse-denoise” loop, which enhances details and ensures global consistency.
Despite its strong performance, naive upsampling in DemoFusion often leads to unintended degradation, which we will discuss in detail in the following section. 

%% file: partial/fig_mean_variance.tex
\begin{figure}[t]
    \centering
    \includegraphics[width=\linewidth]{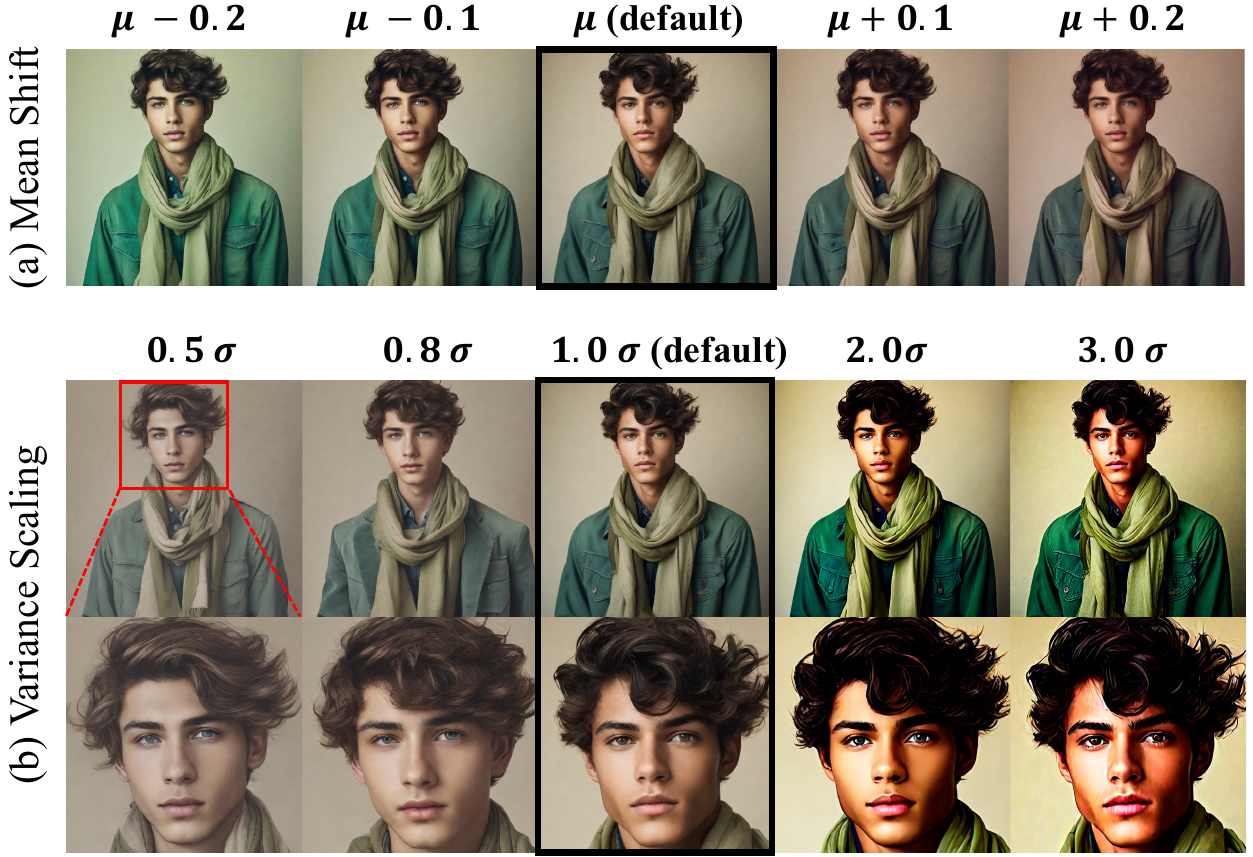}
    \caption{
    \textbf{Influence of latent space mean and variance on decoded output.} 
    We change mean and variance of the latent pixel distribution. (a) Mean shifts in the latent space lead to color shifts in the decoded image. (b) Adjusting latent space variance alters outcomes and changes image frequency characteristics.
    }
    \label{fig:stat_matching}
\end{figure}



%% file: section/3_observations.tex
\section{Key observations}
\label{sec:observations}
Our approach is driven by two key observations.
The first is \textit{patch-level distribution shift}, where statistical variations in the input latents' pixel values affect the characteristics of the output image.
The second is \textit{increased patch monotonicity}, which occurs as the receptive field decreases while the patch size remains fixed.
We identify specific latent space characteristics that influence image quality, and these insights serve as the motivation for our methods.

\subsection{Patch-level distribution shift}
\label{subsec:observation_effect_of_mean_var}

One of our key observations is that the mean and variance of input latents play a significant role in the quality of generated images. 
In LDMs, even small shifts in these statistics can lead to noticeable changes in the output~\cite{hu2024one}. 
As shown in \Cref{fig:stat_matching}, adjusting the mean and variance of latents impacts the quality of decoded images.

\noindent \textbf{Mean shift.} 
As shown in \Cref{fig:stat_matching}(a), shifting the mean of the latent distribution affects the color balance in the reconstructed image, introducing perceptible color shifts. 

\noindent \textbf{Variance scaling.} 
\Cref{fig:stat_matching}(b) demonstrates how scaling the variance impacts the fine details and textures in the generated image. 
Lower variance ($0.5\sigma$) results in a more blurred appearance, while higher variance ($2.0\sigma$) produces excessive contrast in the generated image. 

These findings reveal the importance of matching the mean and variance of dilated patches to those of the initial low-resolution latent.
However, commonly used upsampling methods (e.g., bicubic interpolation) during latent upsampling introduce statistical shifts, as further discussed in Section D.1 of the supplementary.
To address the issue, we propose a normalization technique that statistically aligns the initial low-resolution latent and the dilated patches, mitigating distortion and preserving global coherence.

\input{partial/fig_pixel_redudancy}

\subsection{Increased patch monotonicity}
\label{subsec:observation_pixel_redundancy}
Our second observation relates to pixel redundancy, which varies with receptive field changes caused by image size variation.
In \Cref{fig:pixel_redundancy}, we analyze the pixel redundancy of patches by calculating self-similarity matrices.
As illustrated, the self-similarity matrices show higher mean similarity values in smaller receptive fields (0.57 and 0.70) compared to those in larger receptive fields with an original image (0.54).
This increased redundancy of pixels strengthens low-frequency components, reducing the effect of noise in the diffusion process~\cite{hu2024one}, leading to quality degradation.

This observation is not entirely new; similar insights have been reported in Simple Diffusion~\cite{hoogeboom2023simple}, which notes that pixel redundancy increases as overall image size increases.
While our finding aligns with this observation, it diverges in a significant way: we focus on pixel redundancy within varying receptive fields of a \textit{fixed-resolution} patch, rather than across entire images of \textit{different resolutions}.

To further clarify the distinction, we illustrate the results of applying Simple Diffusion and our APT to DemoFusion, as shown in \Cref{rebut:fig_simplediffusion}.
DemoFusion~\cite{du2024demofusion} itself can be interpreted as following the perspective of Simple Diffusion since it uses standard beta scheduling, given that local patches are sampled to match the pretrained size. 
However, it fails to account for the growing pixel redundancy that comes with increasing image size, resulting in a blurred texture in the grass due to the weakened noise effect during the sampling process.
We also show that naively applying the beta scheduling modification of Simple Diffusion based on image resolution leads to a noise schedule misaligned with patch size, resulting in severe degradation of the output image with unnatural distortions.
In contrast, APT achieves best results, retaining fine local details without introducing unintended distortions.
This suggests a fundamental difference between our proposed method and Simple Diffusion.

\input{partial/rebut_fig_simplediffusion}



%% file: partial/fig_pixel_redudancy.tex
\begin{figure}[t]
    \centering
    \includegraphics[width=\linewidth]{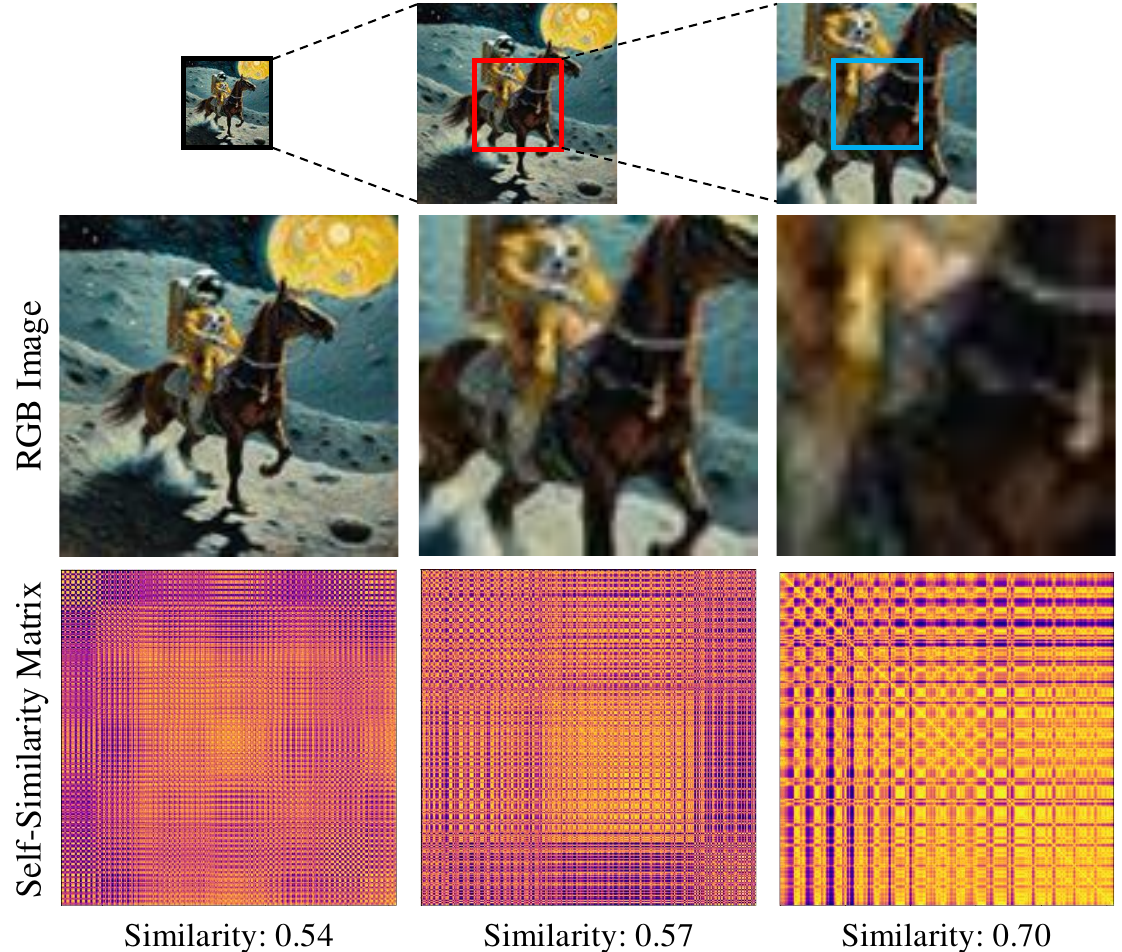}
    \caption{
        \textbf{Comparison of self-similarity (pixel rrdundancy) across different receptive fields.}
        Self-similarity is measured in images at multiple resolutions with a fixed patch size.
        Self-similarity matrices represent pixel-wise similarity, where yellow indicates higher values and purple indicates lower values.
        We calculate similarity by subtracting the average L2 distance between normalized RGB pixel values from 1.
        As the image size increases (left to right), the receptive field of patch decreases and mean similarity increases, indicating stronger pixel redundancy in certain regions.
    }
    \label{fig:pixel_redundancy}
\end{figure}

%% file: partial/rebut_fig_simplediffusion.tex
\begin{figure}[t]
    \centering
    \scriptsize
    \includegraphics[width=1\linewidth]{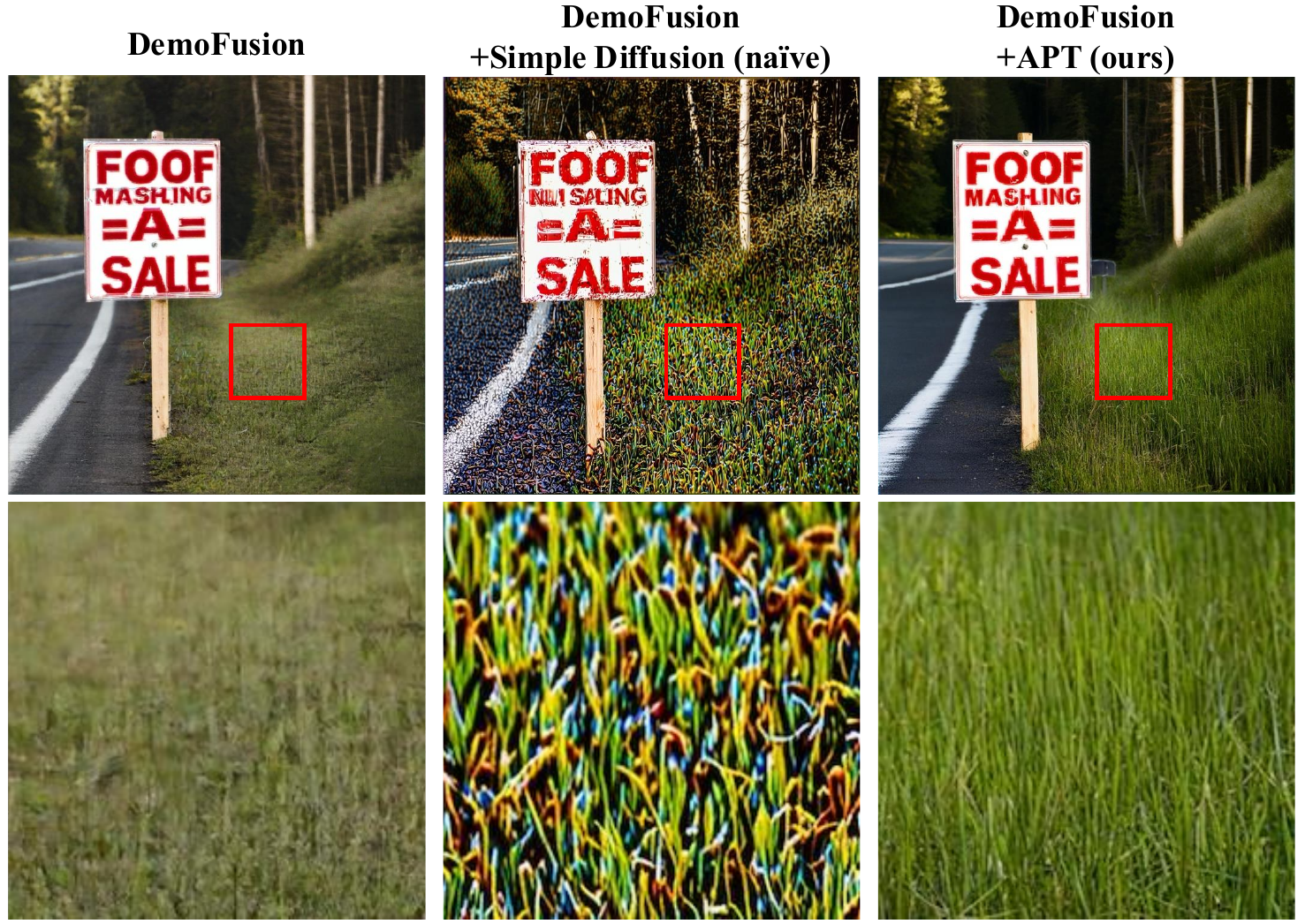}
    \caption{
       \textbf{Simple Diffusion v.s APT}. Qualitative comparison of results applying Simple Diffusion and APT to DemoFusion.
    }
    \label{rebut:fig_simplediffusion}
\end{figure}

%% file: section/4_method.tex
\input{partial/fig_concept}
\section{Methods}
\label{sec:method}

\subsection{Preliminary}
\label{subsec:preliminaries}

Our proposed method is built on LDMs, leveraging an encoder $\mathcal{E}$ and a decoder $\mathcal{D}$ to bridge the data space and the latent space.
In this pipeline, an input image $x\in\mathbb{R}^{h'\times w'\times 3}$ is firstly compressed to a latent $z\in\mathbb{R}^{h\times w\times c}$ where $h < h'$, $w < w'$, and $z=\mathcal{E}(x)$.
Then, the diffusion process (i.e., forward process) is applied by incrementally adding Gaussian noise over $T$ steps for distribution shift from the input latent distribution to standard Gaussian distribution following a Markov chain expressed as:
\begin{equation}
    q(z_t|z_{t-1}) = \mathcal{N}(\sqrt{1-\beta_t}z_{t-1}, \beta_t\mathbf{I}).
    \label{eq:diffuse_latent}
\end{equation}
Here, $z_0=z$ and $z_t$ is the noisy latent at each step $t=1,\dots,T$. $\beta_t$ is a coefficient used in variance scheduling, controlling the noise level. This schedule is typically determined heuristically, considering dataset characteristics such as resolution or diversity~\cite{ho2020denoising, song2020score, karras2022elucidating}.
The diffusion model learns a reverse process to denoise $z_T\sim\mathcal{N}(0,I)$ back to $z_0$, predicting a denoised latent representation. The predicted latent, $\hat{z}_0$, is then decoded into image space as $\hat{x} = \mathcal{D}(\hat{z}_0)$.

\subsection{Adaptive path tracing}
\label{subsec:concept}

\Cref{fig:concept} illustrates the overall concept of APT. In (a), we begin with a pre-trained LDM, where the diffusion process is designed to operate within the manifolds of the original pre-trained resolution, $M_t^{LR}$. The noise schedule and the denoising steps are optimized for this fixed resolution, limiting the ability to scale up to higher resolutions.

In (b), patch-based methods attempt to adapt to higher resolutions by upsampling the initial latent using conventional techniques such as bicubic interpolation. 
However, this approach causes the latent to deviate from the high-resolution target manifold, $M_0^{HR}$.
In addition, using the pre-trained beta scheduling regardless of pixel redundancy causes the successive deviation of the upsampled latent from high-resolution data manifolds.
These misalignments lead to suboptimal denoising steps and quality degradation in the final output.

In (c), our proposed APT addresses these issues through two key techniques.
Statistical Matching is applied to dilated  patches to adjust the mean and variance with the initial latent, aligning it more closely with $M_0^{HR}$. 
This reduces the gap between the upsampled latent and the target manifold, setting a better initial point for the denoising process. 
Scale-aware Scheduling, on the other hand, is applied during local patch sampling to dynamically adjust the noise schedule based on scaling factors. 
This ensures that the upsampled latent maintains the desired SNR throughout the diffusion and denoising steps, guiding the noised latents toward the high-resolution manifolds, $\mathcal{M}_t^{HR}$.
These two strategies enable APT to refine the latent more effectively.
Furthermore, by refining an initial point and adapting step sizes, APT enables a shortcut-sampling, which reduces the number of denoising steps needed to generate high-quality high-resolution images with minimal quality trade-offs.

\subsection{Statistical matching}
\label{subsec:statistical_matching}

By aligning the mean and variance of each dilated patch with those of the reference latent, Statistical Matching corrects their shifts and ensures consistent reconstruction among the patches.
In our case, the input low-resolution latent serves as the reference latent.

Following previous works~\cite{du2024demofusion, lin2025accdiffusion}, dilated patches with stride $S_h=\frac{H}{h}$ and $S_w=\frac{W}{w}$ at timestep $t$ are defined as:
\begin{equation}
    d_{t}^{k} = z^\text{HR}_t[i::S_h,j::S_w,:],
\end{equation}
where $i\in\{0,...,S_h-1\}$ and $j\in\{0,...,S_w-1\}$. The index of dilated patches is represented by $k = i\times S_w + j + 1$ where $k\in\{1,\dots,S_h\times S_w\}$.

While dilated patches exhibit structural similarity with the reference latent, they differ in statistical properties, particularly in their mean and variance.
We will elaborate this with experiments in Figure D in the supplementary material.
To mitigate the distribution shift caused by conventional upsampling and enhance global coherence, we normalize each dilated patch as:
\begin{equation}
    \Tilde{d}_{0}^{k} = \frac{\sigma_{z_0}}{\sigma_{d_{0}^{k}}} \left( d_{0}^{k} - \mu_{d_{0}^{k}} \right) + \mu_{z_0}.
\end{equation} 
Here, $\mu_{z_0}$ and $\sigma_{z_0}$ denote the mean and variance of $z_0$, while $\mu_{d_{0}^{k}}$ and $\sigma_{d_{0}^{k}}$ represent those of $d_{0}^{k}$.

\subsection{Scale-aware scheduling}
\label{subsec:adjusting_beta_scheduling}

At each timestep $t$, the high-resolution latent $z_t^\text{HR}$ is divided into multiple overlapping local patches, with an overlap ratio  $r \in (0, 1)$, defined as:
\begin{equation}
    p_t^l = z_t^\text{HR}[i:i+h, j:j+w, :],
\end{equation}
where  $l \in \{1, \dots, (\frac{H}{hr}-1)(\frac{W}{w r}-1)\}$.

Our observation in \Cref{subsec:observation_pixel_redundancy} indicate that as the receptive field of each patch decreases, the pixel redundancy within each patch $p_0^l$ increases. 
We adjust the intensity of the noise $\beta_t$ in \Cref{eq:diffuse_latent} according to the pixel redundancy to address this, where the primary factor determining pixel redundancy is the upscaling factor $s$, as the patch size remains fixed to the low-resolution latent.
The adjustment is defined as:
\begin{equation}
    \beta_t = \left((\beta_0)^{\eta_s} + t\times \frac{(\beta_T)^{\eta_s} - (\beta_0)^{\eta_s}}{T}\right)^{\frac{1}{\eta_s}},
\end{equation}
where $\beta_T, \beta_0 \in \mathbb{R}$ are pre-defined scalars, and $\eta_s$ is a parameter controlling the growth rate of noise intensity $\beta_t$ depending on the scaling factor $s$.
As $s$ grows, pixel redundancy increases, necessitating faster noise growth to maintain a balanced SNR, which in turn requires a corresponding increase in $\eta_s$.
Increasing $\eta_s$ results in a sharper growth in $\beta_t$, allowing better denoising as illustrated in \Cref{fig:concept}(c). 
We validate our proposed methods in our experiments.

\input{partial/tab_main_comparison}

%% file: partial/fig_concept.tex
\begin{figure*}[htbp]
    \centering
    \includegraphics[width=0.95\textwidth]{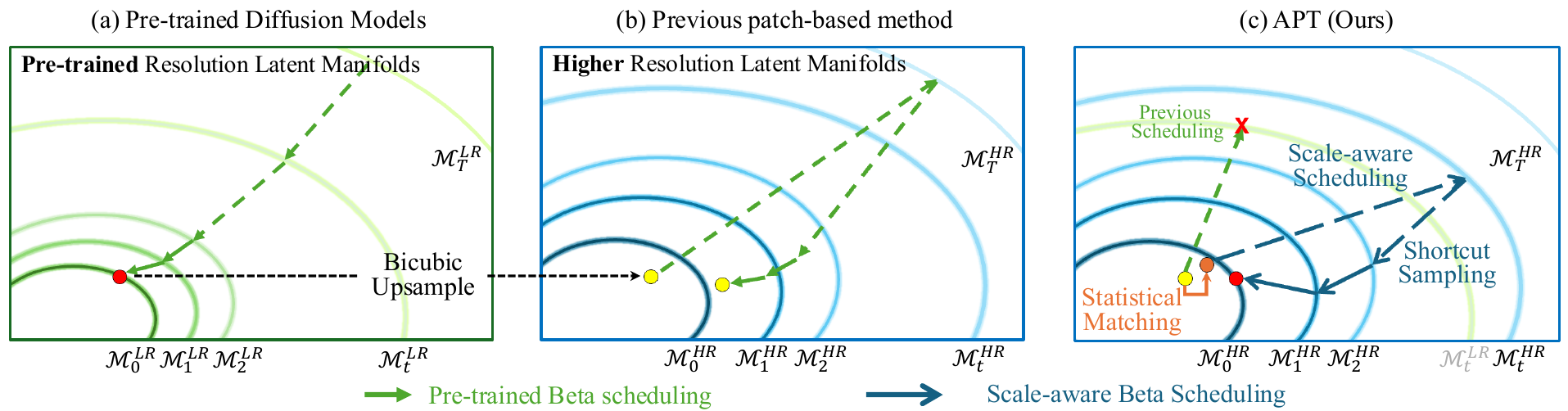}
    \vspace{-0.6em}
    \caption{
        \textbf{Overall concept of APT.}
        (a) In pre-trained latent diffusion models, the diffusion process is tailored to pre-trained resolution (64$^2$ or 128$^2$) latent manifolds.
        (b) Bicubic upsampling shifts the latent representation to a higher resolution manifold, but not perfectly aligned. 
        In addition, previous patch-based methods apply standard beta scheduling, which may not fully adapt to higher resolution needs.
        (c) Our approach, APT, utilizes Statistical Matching to adjust the mean and variance of sampled patches, aligning the latent more closely with the higher resolution manifold, and Scale-aware Scheduling to adapt the diffusion step size for efficient high-resolution latent generation. APT enables shortcut sampling, reducing the number of denoising steps required to generate high-quality.
    }
    \label{fig:concept}
    \vspace{-1em}
\end{figure*}

%% file: partial/tab_main_comparison.tex
\renewcommand{\arraystretch}{1.1}
\begin{table*}[]
\centering
\vspace{-0.8em}
\resizebox{\textwidth}{!}{
\begin{tabular}{c|cc|cc|c|cc|cc|c}
\toprule
\multirow{2}{*}{Method} & \multicolumn{5}{c|}{2048$\times$2048 ($\times$4)} & \multicolumn{5}{c}{4096$\times$4096 ($\times$16)} \\ \cline{2-11} 
                        & MUSIQ $\uparrow$ & CLIPIQA $\uparrow$ & $\text{FID}_{256} \downarrow$ & $\text{KID}_{256} \downarrow$ & Time & MUSIQ $\uparrow$ & CLIPIQA $\uparrow$ & $\text{FID}_{256} \downarrow$ & $\text{KID}_{256} \downarrow$ & Time \\ 
\hline\midrule
SDXL Direct Inference   & 58.1   & 0.585 & 57.7 & 0.0297 & 55 sec & 33.7  & 0.549 & 86.6 & 0.0489 & 13 min \\
ScaleCrafter            & 60.8   & 0.619 & 35.5  & 0.0103 & 63 sec & 38.0  & 0.530 & 54.9 & 0.0103 & 22 min \\
FouriScale              & 56.0   & 0.584 & 53.2  & 0.0183 & 127 sec & 31.8 & 0.515 & 77.0 & 0.0260 & -- \\
HiDiffuion              & 59.9 & 0.607 & 38.0   & 0.0114 & 39 sec & 39.9  & 0.554 & 127.4   & 0.0787 & 4 min \\
\hline
DemoFusion              & 56.6   & 0.587 & 42.5  & 0.0211 & 168 sec & 38.9 & 0.548 & 33.6  & 0.0117 & 22 min \\
$\text{DemoFusion}^\dagger$       & 53.3 (-5.8\%)  & 0.545 (-7.2\%) & 46.3 (+8.9\%)  & 0.0231 (+0.9\%) & 106 sec & 36.9 (-0.5\%)  & 0.517 (-5.7\%) & 37.2 (+10.7\%)  & 0.0133 (+13.7\%) & 13 min \\
$\text{DemoFusion}^\dagger$\textbf{+APT} & 59.0 \textbf{(+4.2\%)} & 0.632 \textbf{(+7.7\%)} & 37.1 \textbf{(-12.7\%)}  & 0.0160 \textbf{(-24.1\%)} & 106 sec & 40.3 \textbf{(+3.6\%)} & 0.598 \textbf{(+0.9\%)} & 31.5 \textbf{(-6.3\%)}  & 0.0104 \textbf{(-11.1\%)} & 13 min \\ 
\cdashline{1-11}
AccDiffusion            & 56.9   & 0.569 & 36.5  & 0.0174 & 173 sec & 38.7 & 0.536 & 33.7  & 0.0113 & 23 min \\
$\text{AccDiffusion}^\dagger$     & 50.5 (-11.4\%)  & 0.516 (-9.3\%) & 46.1 (+26.3\%)  & 0.0230 (+32.2\%) & 109 sec & 36.5 (-5.7\%)  & 0.485 (-9.5\%) & 38.0 (+11.1\%)  & 0.0127 \textbf{(+12.4\%)} & 14 min \\
$\text{AccDiffusion}^\dagger$\textbf{+APT} & 56.6 \textbf{(-0.5\%)} & 0.595 \textbf{(+4.6\%)} & 37.6 \textbf{(+3.0\%)}  & 0.0175 \textbf{(+0.0\%)} & 109 sec & 40.3 \textbf{(+4.1\%)} & 0.557 \textbf{(+3.9\%)} & 33.8 \textbf{(+0.3\%)}  & 0.0131 (+14.9\%) & 14 min \\ 
\bottomrule
\end{tabular}}
\vspace{-0.5em}
\caption{
    \textbf{Quantitative comparison results}, with various models at resolutions of 2048×2048 and 4096×4096. MUSIQ and CLIPIQA assess perceptual quality and semantic alignment, while $\text{FID}_{256}$ and $\text{KID}_{256}$ measure fine detail quality. Inference time is also included. $\dagger$ indicates that models perform shortcut sampling with 30 steps  compared to the baseline’s 50 steps. The inference time of FouriScale is not evaluated at $\times$16 due to out-of-memory on our A5000 GPU.
}
\label{tab:main_comparison}
\end{table*}

%% file: section/5_experiments.tex
\section{Experiments}
\label{sec:experiments}


\subsection{Experimental settings}
\label{sec:settings}

\noindent{\textbf{Models.}}
To evaluate APT’s effectiveness in improving the ``upsample-diffuse-denoise" loop of patch-based methods, we integrate APT into DemoFusion~\cite{du2024demofusion} and AccDiffusion~\cite{lin2025accdiffusion}, and compare their performance against the baselines.
We also validate our models using several training-free methods, including SDXL~\cite{podellsdxl}, ScaleCrafter~\cite{he2023scalecrafter}, FouriScale~\cite{huang2024fouriscale}, and HiDiffusion~\cite{zhang2023hidiffusion}, which propose alternative patch-based solutions. 
For fair comparison, we use SDXL as a base diffusion model for all models.

\noindent\textbf{Dataset.} Since the commonly used benchmarks such as COCO~\cite{lin2014microsoft} and LAION~\cite{schuhmann2022laion} have lower-resolution images (often below 1K), they are inadequate for evaluating high-resolution image quality. 
Hence, we construct an image-caption paired test set with 1K randomly sampled images from OpenImages~\cite{kuznetsova2020open}, all with resolutions exceeding 3K.
The image captions are generated using BLIP2~\cite{li2023blip}. 

\noindent\textbf{Metrics.} We employ MUSIQ~\cite{ke2021musiq},  CLIPIQA~\cite{wang2023exploring}, $\text{FID}_c$ and $\text{KID}_c$ for quantitative evaluation metrics.
MUSIQ and CLIPIQA are NRIQA metrics aligned with human preferences to evaluate the overall image quality~\cite{yue2023resshift}.
$\text{FID}_c$ and $\text{KID}_c$ focus on fine details by analyzing \textit{cropped} patches following Chai et al~\cite{chai2022any}.
We would like to note that we do not utilize FID~\cite{heusel2017gans} and KID~\cite{binkowski2018demystifying} metrics since they are inadequate for measuring high resolution image quality. This has already been highlighted in previous studies~\cite{podellsdxl, kirstain2023pick} and we provide a more detailed discussion in the supplementary material. 
Additionally, we measure inference time to validate efficiency using a NVIDIA RTX A5000.

\input{partial/fig_main_comparison}

\subsection{Quantitative results}
\label{sec:quantitative}
We provide a set of quantitative experiments to evaluate the effect of APT, as shown in \Cref{tab:main_comparison}. The models are broadly divided into two categories: non-patch-based (top 4 models) and patch-based (bottom 6 models) approaches.

In the non-patch-based group, HiDiffusion demonstrates relatively fast inference time. However, its performance is inconsistent, showing a marked decline as the target resolution increases. This shortcoming becomes even more evident in the following qualitative analysis. Similarly, while ScaleCrafter performs well at 2K resolution, it suffers a significant drop in all metrics as the resolution grows larger.

In the patch-based group, we demonstrate APT’s effectiveness through two widely used baselines, DemoFusion and AccDiffusion. We compare the performance of their naive shortcut versions- 
reducing the denoising steps from 50 to 30—with the versions of using APT. While naive timestep reductions accelerate inference time, they result in considerable performance degradation across all metrics. By contrast, APT not only retains the inference speed gains (around 40\% faster), but also mitigates the performance drop, even surpassing the original baselines in many cases.


\subsection{Qualitative results}
\label{sec:qualitative}

In \Cref{fig:qualitative_comparison}, we analyze both object-centric and landscape-oriented scenes.
We include HiDiffusion as a non-patch-based model. While it maintains overall performance at 2K resolution, it faces scalability limitations and generates severe artifacts at 4K resolution.
DemoFusion and AccDiffusion effectively capture both global content and fine details at all resolutions. However, the aforementioned two key issues in these frameworks introduce distortions, such as blurry necklaces and unnatural textures on veranda handrails. By integrating APT, these limitations are mitigated. Both DemoFusion+APT and AccDiffusion+APT enhance fine-grained details in content and textures.

\subsection{Ablation study}
\label{sec:ablation}

\subsubsection{Individual components}
To assess the individual contributions of Statistical Matching (SM) and Scale-aware Scheduling (SaS), we evaluate models incorporating these components into DemoFusion, as shown in \Cref{tab:ablation}. Each component produces clear improvements across all metrics, with the combined APT model delivering the best overall performance, surpassing the original DemoFusion while also reducing inference time. Qualitative results in \Cref{fig:intra_ablation} further highlight the effectiveness of each technique and their compatibility. Additionally, APT performs well on a different baseline, AccDiffusion, as demonstrated in \Cref{fig:inter_ablation}.


\subsubsection{Correlation of $\eta$ and scaling factor}
To examine the relationship between the scheduling parameter $\eta$ and scaling factor, we conducted an ablation study, as visualized in \Cref{fig:experiments_beta_scheduling}.
This study explores our hypothesis from \Cref{subsec:adjusting_beta_scheduling}, which suggests that the optimal $\eta$ value should correlate with the scaling factor, \textit{increasing} noise intensity to address \textit{increased} pixel redundancy at higher resolutions. 
Our results confirm this hypothesis, showing that as the scaling factor increases, the optimal $\eta$ value also rises, achieving the best performance for each resolution. 
This adaptability allows APT to dynamically adjust the noise schedule, maintaining an optimal signal-to-noise ratio (SNR) across scales, which is essential for preserving high image quality across varying resolution levels.

\input{partial/tab_intra_ablation}
\input{partial/fig_intra_ablation}

\subsubsection{Impact of crop size in patch based metric}

As shown in \Cref{rebut:fig_timestep}(a), we evaluate models with varying the patch size from $1024^2$ to $256^2$.  
In terms of FID$_{1024}$, DemoFusion+APT and DemoFusion yield similar global quality.
However, as the crop size decreases and finer details are emphasized, the differences become more apparent.
This trend suggests that our method not only effectively refines details but also maintains global coherence.

\subsubsection{Shortcut timesteps}
We present \Cref{rebut:fig_timestep}(b), which quantitatively compares different methods across various shortcut timesteps.
Both DemoFusion and DemoFusion+APT show improved performance as timesteps increase.
However, beyond 30 timesteps, the improvements become marginal. 
Given this trend, we conclude that 30 timesteps provide an optimal trade-off between efficiency and performance.

\input{partial/fig_inter_ablation}
\input{partial/fig_beta_searching}

\input{partial/rebut_fig_timestep}

%% file: partial/fig_main_comparison.tex
\begin{figure*}[htbp]
    \centering
    \includegraphics[width=0.98\textwidth]{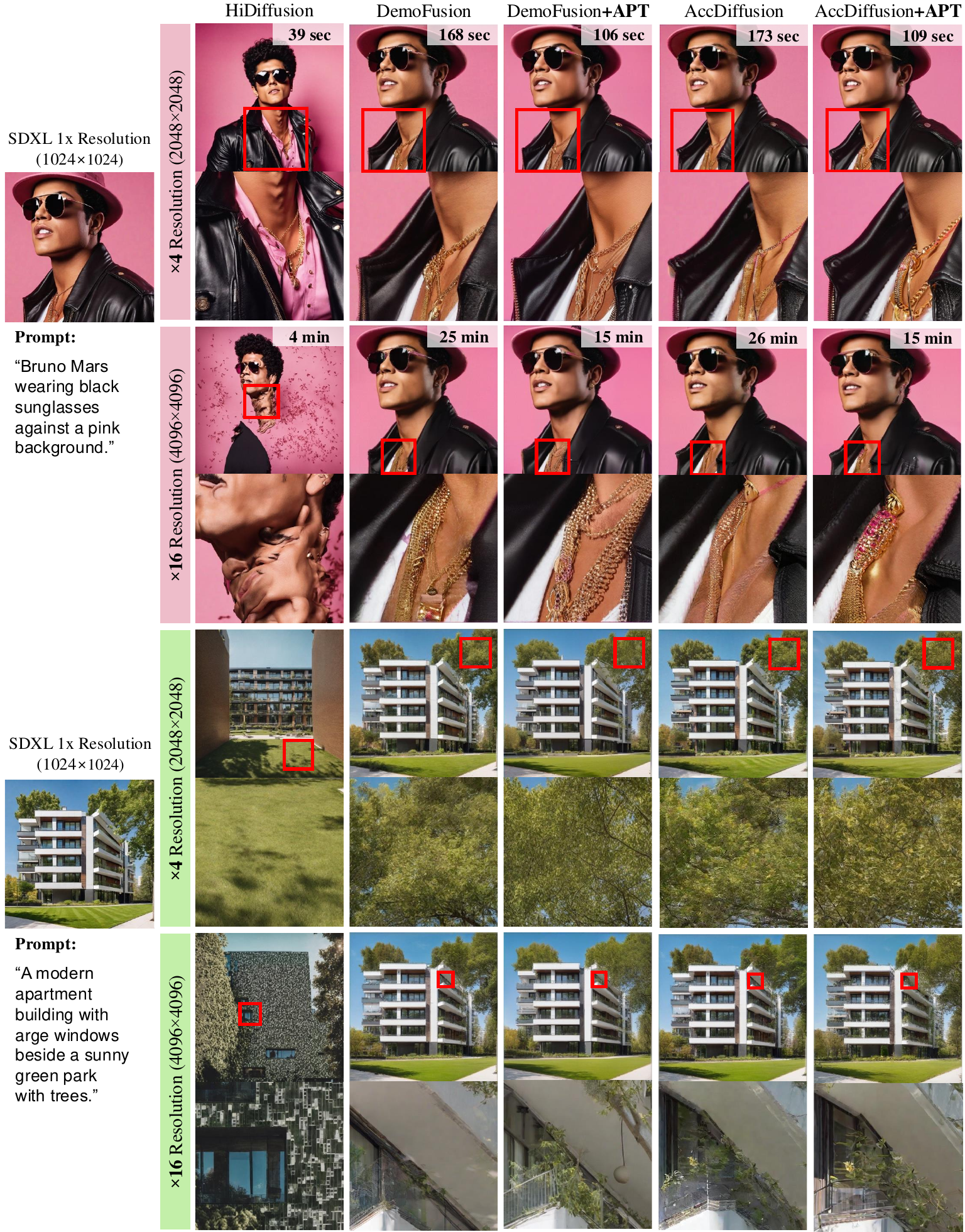}
    \vspace{-0.6em}
    \caption{
        \textbf{Qualitative comparison. }
        Visual comparison of high-resolution generations across multiple methods for $\times4$ and $\times16$ scales. 
        It is recommended to zoom in to examine fine details and differences in image fidelity.
    }
    \label{fig:qualitative_comparison}
\end{figure*}

%% file: partial/tab_intra_ablation.tex
\begin{table}[]
\centering
\vspace{-0.5em}
\renewcommand{\arraystretch}{1.3}
\resizebox{\linewidth}{!}{%

    \begin{tabular}{lccccc}
    \toprule
               & MUSIQ$\uparrow$ & CLIPIQA$\uparrow$ & $\text{FID}_{256}$$\downarrow$ & $\text{KID}_{256}$$\downarrow$ & Time \\ \hline
               \hline
    DemoFusion \cite{du2024demofusion} (50/50)   & 43.42  & 0.549 & 40.72 & 0.019 & 11 min \\
    DemoFusion (30/50)   & 38.32 & 0.476 & 49.10 & 0.022 & 6 min \\
    + SM only  & 43.83 & 0.557 & 38.92 & 0.016 & 6 min \\
    + SaS only  & 45.55 & 0.587 & 40.80 & 0.018 & 6 min \\
    + SM + SaS  ($=$APT)  & \textbf{46.54} & \textbf{0.606} & \textbf{38.24} & \textbf{0.007} & 6 min \\
    \bottomrule
    \end{tabular}
}%
\caption{
    \textbf{Ablation study on APT components.} 
    SM indicates Statistics Matching, and SaS indicates Scale-aware Scheduling. 
}
\label{tab:ablation}
\end{table}

%% file: partial/fig_intra_ablation.tex
\begin{figure}[t]
    \centering
    \includegraphics[width=\linewidth]{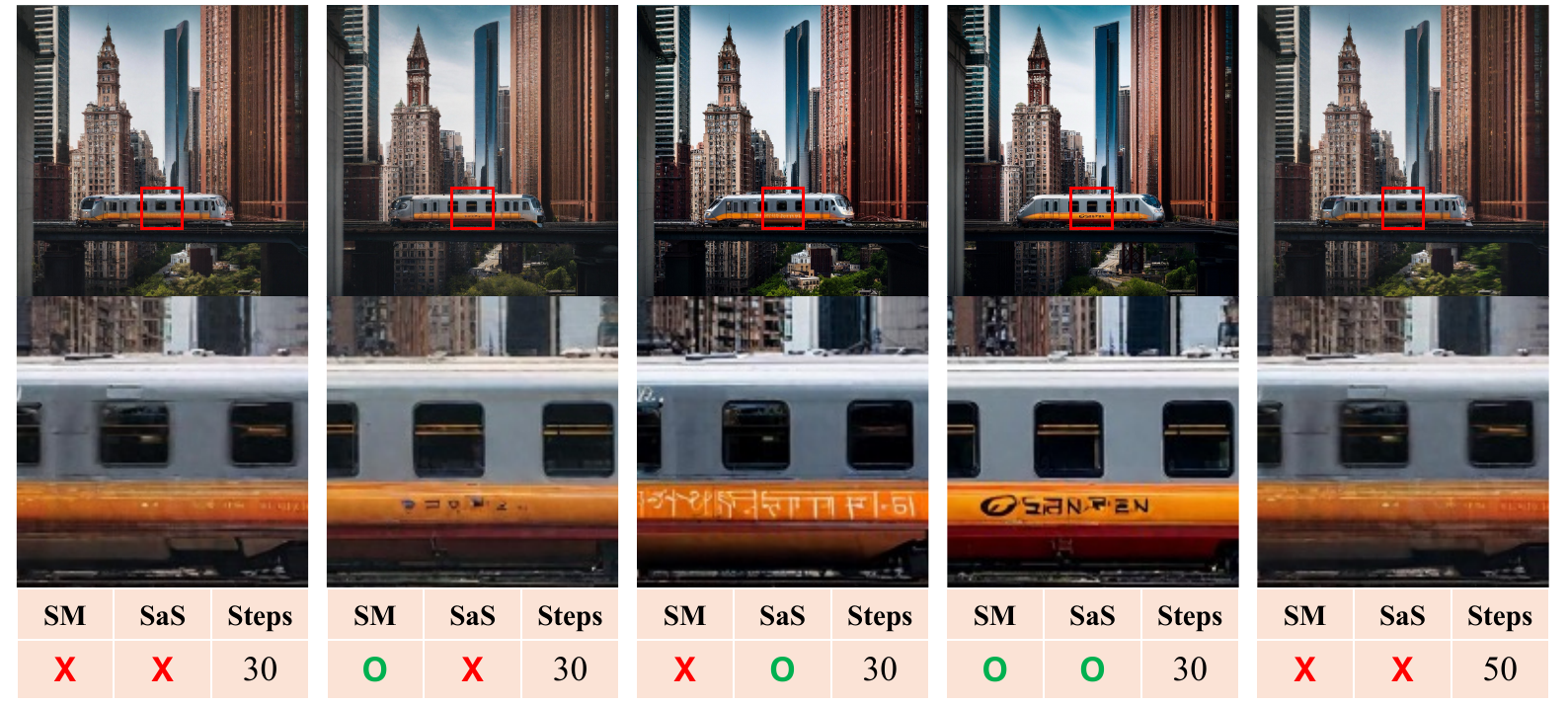}
    \caption{
        \textbf{Qualitative results of ablation study on APT components.}
        Effect of each component in our method (SM and SaS) compared with DemoFusion. 
        The presence of each method is indicated below each image, showing its impact on image quality.
    }
    \label{fig:intra_ablation}
\end{figure}



%% file: partial/fig_inter_ablation.tex
\begin{figure}[t]
    \centering
    \includegraphics[width=\linewidth]{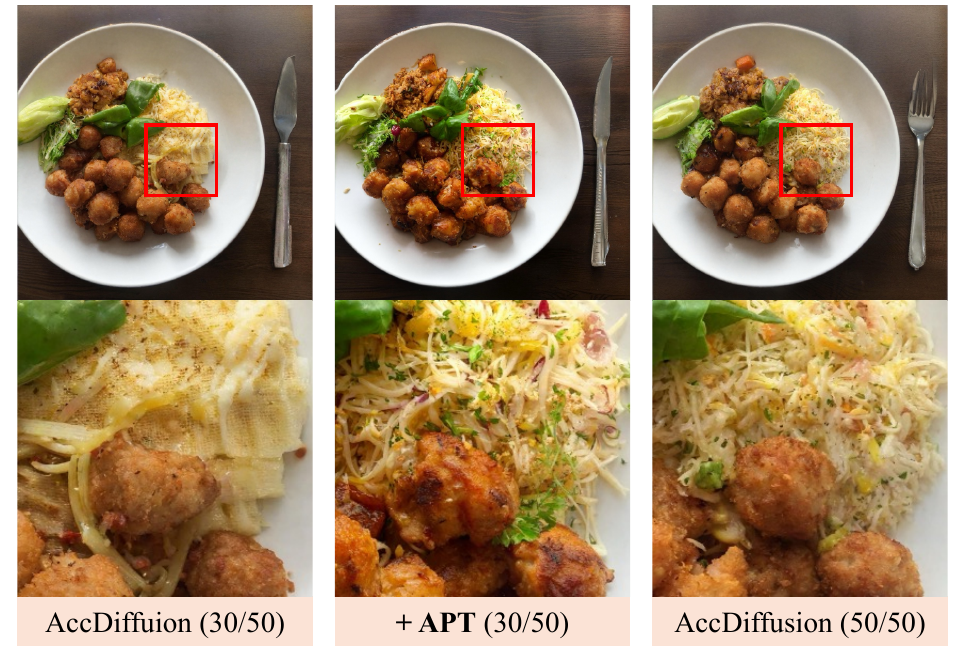}
    \caption{
        \textbf{Qualitative results of APT applied to AccDiffusion.} 
    }
    \label{fig:inter_ablation}
\end{figure}



%% file: partial/fig_beta_searching.tex
\begin{figure}[t]
    \centering
    \includegraphics[width=\linewidth]{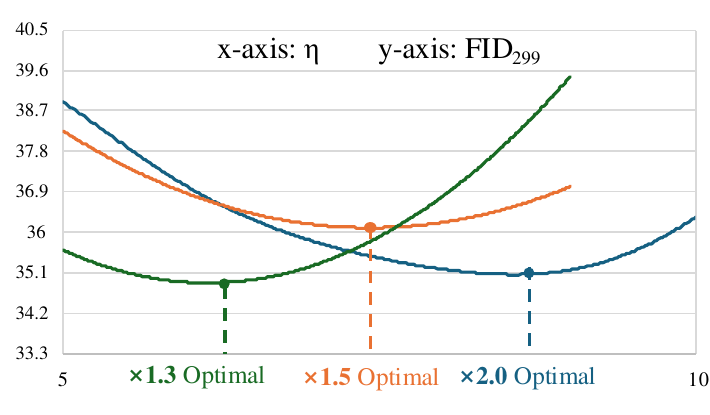}
    \caption{
       \textbf{Experiments on the relationship between scheduling parameter $\eta$ and scaling factor.} Results show that optimal $\eta$ increases with scaling, confirming the need for dynamic noise scheduling to maintain image quality across resolutions.
    }
    \label{fig:experiments_beta_scheduling}
\end{figure}

%% file: partial/rebut_fig_timestep.tex
\begin{figure}[t]
    \centering
    \includegraphics[width=1\linewidth]{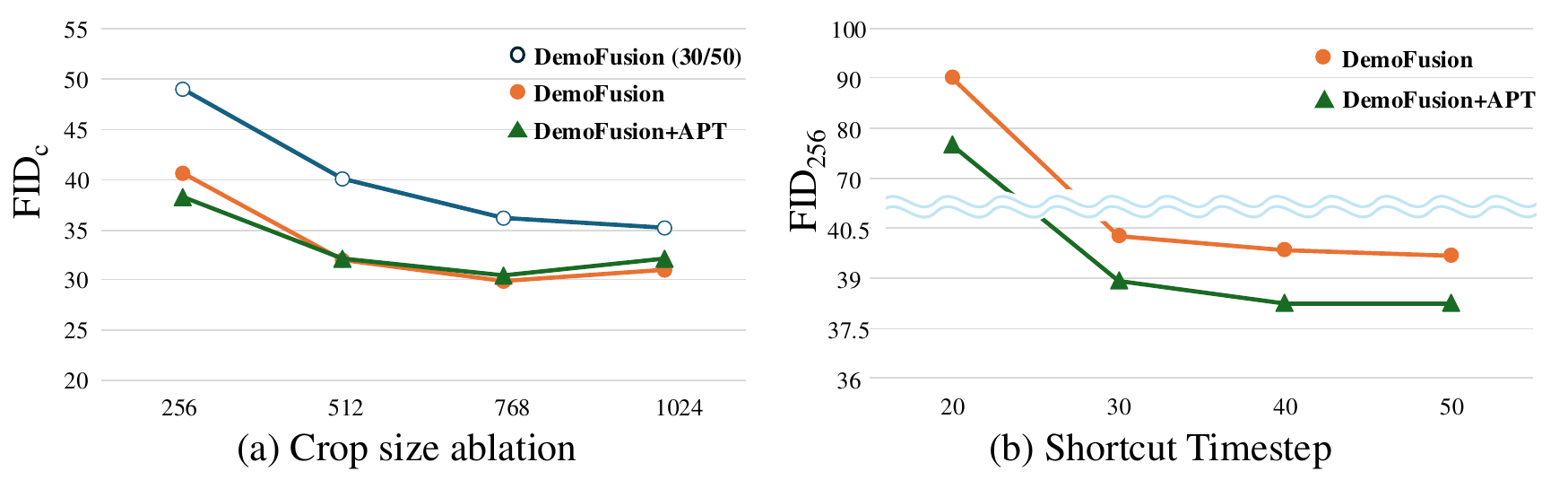}
    \caption{
       \textbf{Ablation study on crop size and shortcut timesteps} in patch-based metrics and performance at resolution 3K.
    }
    \label{rebut:fig_timestep}
\end{figure}

%% file: section/6_conclusion.tex
\section{Conclusion}
\label{sec:conclusion}


We introduce APT (Adaptive Path Tracing), an effective method for high-resolution image generation within latent diffusion models.
We found that conventional upsampling alters the mean and variance of latents, resulting in unintended image distortions, and increased pixel redundancy in fixed-size patches disrupts pre-trained SNR in diffusion process.
With Statistical Matching and Scale-aware Scheduling, APT addresses these issues, optimizing noise control across scales. 
APT also enables shortcut sampling to improve inference speed without sacrificing quality. 
We hope APT offers a practical solution for high-resolution image generation.

%% file: supple.tex
\clearpage
\renewcommand{\thesection}{\Alph{section}}
\renewcommand{\thefigure}{\Alph{figure}}
\renewcommand{\thetable}{\Alph{table}}
\setcounter{section}{0}
\setcounter{table}{0}
\setcounter{figure}{0}

\setcounter{page}{1}
\maketitlesupplementary

\input{section/A_Limitation}
\input{section/B_Experimental_Detail}

\input{section/C_More_ablations}
\input{section/D_Interpolation_upsampling}
\input{section/E_SimpleDiffusion}
\input{section/F_FID}
\input{section/G_Extreme_Resolution}

%% file: section/A_Limitation.tex
\section{Limitations and future works}
\label{supp_sec:limitation}

While APT introduces significant advancements in high-resolution image generation, it has several limitations that warrant further exploration.
(1) Although APT reduces sampling time by approximately 40\%, the overall inference speed remains a bottleneck for real-time or large-scale applications, particularly when generating ultra-high-resolution images.
(2) As a training-free framework, APT relies on the capabilities of the backbone diffusion model and thus cannot generate patch-level images that surpass the inherent quality of the pre-trained model.
(3) Despite improvements in image fidelity and fine details, APT, like previous methods, still encounters issues with small object repetition in complex or highly repetitive scenes.

To overcome these limitations, future work could focus on optimizing the number of patches or resolving the needs of the progressive upsampling process to further enhance efficiency without compromising quality. Additionally, integrating lightweight learning mechanisms or adaptive refinement techniques during inference could help address patch-level quality dependencies and mitigate repetitive artifacts. Exploring hybrid methods that combine training-free approaches with minimal fine-tuning may also provide a promising avenue for scaling APT to real-time applications and improving robustness in challenging image scenarios.


%% file: section/B_Experimental_Detail.tex
\section{Experimental details}
\label{supp_sec:experiment_detail}

\subsection{Dataset}

\textbf{Negative prompts.} We utilize a fixed negative prompt, ``blurry, ugly, duplicate, poorly drawn, deformed, mosaic" from~\cite{du2024demofusion}, across all comparison models to ensure high-quality, high-resolution image generation. 

\noindent\textbf{Test set.}
We construct a test set for main comparison shown in Table 1. in the main paper. The test set consists of an image-caption paired dataset with 1,000 randomly sampled images from OpenImages~\cite{kuznetsova2020open}, all with resolutions exceeding 3K. In order to match a 1:1 resolution ratio of test images, we crop the original image at the center based on the shorter side length. Captions for the cropped images are generated using BLIP2~\cite{li2023blip}.

\noindent\textbf{Validation set.}
We construct a validation set and use it for ablation studies to efficiently evaluate our proposed methods. Similar to the test set, the validation set consists of an image-caption paired dataset with 400 randomly sampled images from OpenImages~\cite{kuznetsova2020open}, all with resolutions exceeding 3K.
To ensure fairness, images in the validation set are not included in the test set, as hyperparameter tuning (e.g., $\eta$ and shortcut timesteps) based on the validation set could introduce bias. 

We conduct ablation studies on the validation set to evaluate the effectiveness of APT components, including the optimal selection of $\eta$ for different scaling factors and an analysis of shortcut timestep configurations. For efficiency and reliability, all validations are performed with images generated at 3K resolution.

Our proposed method, APT, consistently improves DemoFusion~\cite{du2024demofusion} across both the test and validation sets. These results suggest that scale-adaptive scheduling and statistical matching contribute to stable performance improvements in high-resolution image generation.

\subsection{Patch-based metrics}

To evaluate the fine details of generated high-resolution images, we utilize FID$_c$ and KID$_c$~\cite{chai2022any}, where scores are computed on smaller cropped patches rather than globally resized images at 299$\times$299 resolution. 

For more details, patches are cropped to $256 \times 256$, ensuring compatibility with FID~\cite{heusel2017gans} and KID~\cite{binkowski2018demystifying} computation. 
As shown in Figure 10(a) in the main paper, the model performance trend remains stable across different crop sizes.
Crop locations are randomly determined but fixed across the ground truth, corresponding baseline- and our results for fair comparison. 
We utilize 50,000 patches to calculate FID$_c$ and KID$_c$, providing a robust metric for assessing fine details in high-resolution image generation.

\subsection{Implementation details}

We utilize SDXL~\cite{podellsdxl} and DDIM~\cite{songdenoising} scheduler for experiments. The pseudocode of APT for DemoFusion is shown in \Cref{alg:pseudocode}.

\input{partial/pseudocode}

%% file: partial/pseudocode.tex
\begin{algorithm*}[htbp]
\small
\caption{High resolution Image Generation Process of APT for DemoFusion}

\textbf{Input}: $h, w$ \Comment{Pre-trained resolution} \\
\indent \quad \quad \quad 
$h^\text{hr}, w^\text{hr}$ \Comment{Target higher resolution} \\
\indent \quad \quad \quad
$\theta, \mathcal{D}$ \Comment{Pre-trained Stable diffusion model parameter and Decoder} \\
\indent \quad \quad \quad
$\Phi, y$ \Comment{Conventional upsampler and Prompt} \\
\indent \quad \quad \quad
$T_0, \eta$ \Comment{Shortcut timestep and Control parameter for Scale-aware Scheduling} \\
\indent \quad \quad \quad
$\lambda_1, \lambda_2$ \Comment{Decreasing factors from 1 to 0 via Cosine schedule}

\begin{algorithmic}[1]
\State \texttt{\#\#\#\#\#\#\#\#\#\#\#\#\#\#\#\#\#\#\#\#\#\#\#\#\# Phase $1$ : Reference image generation \#\#\#\#\#\#\#\#\#\#\#\#\#\#\#\#\#\#\#\#\#\#\#\#\#}
\State $z^1_T \sim \mathcal{N}(0, I)$ \Comment{Random Initialization}

\For{$t=0$ to $T$}
        \State $\beta_t = \left((\beta_0)^{\eta_1} + t\times \frac{(\beta_T)^{\eta_1} - (\beta_0)^{\eta_1}}{T}\right)^{\frac{1}{\eta_1}}$ \Comment{Pre-trained beta scheduling}
        \State $\alpha_t=1-\beta_t$
    \EndFor
\For{$t = T$ to $1$}
    \State $z^1_{t-1}=\sqrt{\frac{\alpha_{t-1}}{\alpha_t}}z^1_t + \left(\sqrt{\frac{1}{\alpha_{t-1}}-1} - \sqrt{\frac{1}{\alpha_{t}}-1}\right)\cdot\epsilon_\theta(z^1_t,t,y)$ \Comment{Denoising Step}
\EndFor
\State $\mu_{z^1_0}=\mathbb{E}[z^1_0]$ \Comment{\textbf{Mean of reference latent}}
\State $\sigma_{z^1_0}^2=\mathrm{Var}[z^1_0]$ \Comment{\textbf{Variance of reference latent}}
\State $S=\frac{h^\text{hr}}{h}\times\frac{w^\text{hr}}{w}$ \Comment{Progressive upsampling Iteration}
\State \texttt{\#\#\#\#\#\#\#\#\#\#\#\#\#\#\#\#\#\#\#\#\# Phase $2$ : Higher resolution image generation \#\#\#\#\#\#\#\#\#\#\#\#\#\#\#\#\#\#\#\#\#}
\For{$s = 2$ to $S$}
    \State $z'^s_0 = \Phi(z^{s-1}_0, (h\times s, w\times s))$ \Comment{Conventional upsampling}
    \State $L=\frac{1}{r^2}\times\frac{h\times s}{h}\times\frac{w\times s}{w}$
    \State $K=s^2$
    \State $\{d^k_0\}^K_{k=1}=\text{Sampling}_\text{global}(z'^s_0)$ \Comment{Dilated Sampling from DemoFusion}
    \For{$d^k_0$ in $\{d^k_0\}^K_{k=1}$}
        \State $\mu_{d^k_0}=\mathbb{E}[d^k_0]$ 
        \State $\sigma_{d^k_0}^2=\mathrm{Var}[d^k_0]$ 
        \State $\Tilde{d}_{0}^{k} = \frac{\sigma_{z^1_0}}{\sigma_{d_{0}^{k}}} \left( d_{0}^{k} - \mu_{d_{0}^{k}} \right) + \mu_{z^1_0}$ \Comment{\textbf{Normalization for Statistical Matching}}
    \EndFor
    \State $\Tilde{z}^s_0=\text{Fusing}(\{\Tilde{d}^k_0\}^K_{k=1})$
    \For{$t=1$ to $T$}
        \State $\hat{\beta_t} = \left((\beta_0)^{\eta_s} + t\times \frac{(\beta_T)^{\eta_s} - (\beta_0)^{\eta_s}}{T}\right)^{\frac{1}{\eta_s}}$ \Comment{\textbf{Scale-aware Scheduling}}
        \State $\hat{\alpha_t}=1-\hat{\beta_t}$
    \EndFor
    \For{$t = 1$ to $T_0$}
        \State $\Tilde{z}^s_t\sim q(\Tilde{z}^s_t|\Tilde{z}^s_{t-1})$ \Comment{\textbf{Diffusion until Shortcut timestep}}
    \EndFor
    \State $z^s_{T_0}=\Tilde{z}^s_{T_0}$
    \For{$t = T_0$ to $1$}
        \State $\hat{z}^s_t = \lambda_1 \times \Tilde{z}^s_t + (1-\lambda_1) \times z^s_t$ \Comment{Skip Residual}
        \State $\{p^l_t\}^L_{l=1} = \text{Sampling}_\text{local}(\hat{z}^s_t)$ \Comment{Local patch cropping from MultiDiffusion}
        \For{$p^l_t$ in $\{p^l_t\}^L_{l=1}$}
            \State $p^l_{t-1}=\sqrt{\frac{\hat{\alpha}_{t-1}}{\hat{\alpha}_t}}p^l_t + \left(\sqrt{\frac{1}{\hat{\alpha}_{t-1}}-1} - \sqrt{\frac{1}{\hat{\alpha}_{t}}-1}\right)\cdot\epsilon_\theta(p^l_t,t,y)$
            \\ \Comment{Local path Denoising Step From MultiDiffusion with \textbf{Scale-aware Scheduling}}
        \EndFor
        \State $\{d^k_t\}^K_{k=1}=\text{Sampling}_\text{local}(\hat{z}^s_t)$ \Comment{Dilated patch sampling from DemoFusion}
        \For{$d^k_t$ in $\{d^k_t\}^K_{k=1}$}
            \State $d^k_{t-1}=\sqrt{\frac{\alpha_{t-1}}{\alpha_t}}d^k_t + \left(\sqrt{\frac{1}{\alpha_{t-1}}-1} - \sqrt{\frac{1}{\alpha_{t}}-1}\right)\cdot\epsilon_\theta(d^k_t,t,y)$ \Comment{Dilated patch Denoising step from DemoFusion}
        \EndFor
        \State $z^s_{t-1}= \lambda_2 \times \text{Fusing}(\{d^k_{t-1}\}^K_{k=1}) + (1 - \lambda_2) \times \text{Fusing}(\{p^l_{t-1}\}^L_{l=1})$ \Comment{Fusing Local patches and Dilated patches}
    \EndFor
\EndFor
\State \textbf{return} $\mathbf{x}^S_0=\mathcal{D}(z^S_0)$ 
\Comment{Decoding to Image}
\end{algorithmic}
\label{alg:pseudocode}
\end{algorithm*}

%% file: section/C_More_ablations.tex
\section{Additional experiments}
\label{supp_sec:SM_ablation}

To validate the effectiveness of Statistical Matching and Scale-aware Scheduling, we conduct extensive ablation studies using the validation set of 400 image-caption pairs. 

\subsection{Ablation study on statistical matching}

To verify the effectiveness of SM, we additionally leverage another upsamling method, nearest neighbor (NN). 
Since NN copies the pixel values of reference latent for upscaling, the dilated patches of the upsampled latent are exactly same with reference latent which means that the mean and variance of dilated patches from NN upsampled latent are same with those of reference latent.
We compare the shortcut sampling results with bicubic upsampling (DemoFusion), SM after bicubic upsampling and NN upsampling based on DemoFusion~\cite{du2024demofusion} framework.

As shown in \Cref{sup_tab:sm_ablation}, not only SM after bicubic upsampling shows notable improvements over naive bicubic upsampling in FID$_{256}$ and KID$_{256}$, but also SM by NN upsampling achieves better scores than naive bicubic upsampling. 
This improvement suggests that aligning the statistical factors of $d_0^k$ with the reference latent contributes to better initialization for $z_0^\text{HR}$, which plays a crucial role in the diffusion model, as shown in Section D.2. in the main paper.

\input{partial/sup_tab_statistical_matching}



\subsection{Ablation study on scale-aware scheduling}

\input{partial/sup_tab_beta_searching}

\input{partial/sup_fig_snr}

To investigate the relationship between upsample scale and optimal beta scheduling, we conduct an ablation study focused on the parameter $\eta$, which controls noise scheduling in the diffusion process as shown in \Cref{supfig:snr}. Since APT follows the ``upscale-diffuse-denoise" loop~\cite{du2024demofusion}, we empirically search for optimal $\eta$ values across intermediate upscaling steps: $2.0\times$ (1024 $\to$ 2048) $1.5\times$ (2048 $\to$ 3072), and $1.3\times$ (3072 $\to$ 4096). 

The results in \Cref{sup_tab:beta_searching} evaluate $\text{FID}_{299}$, and $\text{KID}_{299}$, with a particular focus on $\text{FID}_{299}$ and $\text{KID}_{299}$ that capture fine-detail quality.
From the results, the optimal $\eta$ values are found to be 2 for $2\times$, 3 for $1.5\times$, and 3.5 for $1.3\times$. 
These values suggest that as the upsampling scale decreases, a slower SNR decay becomes more effective, balancing detail preservation and stability.

Furthermore, as shown in \Cref{sup_fig:beta_search_ablation}, our findings also contribute to qualitative enhancement beyond the noise and detail trade-off. 
This study underscores the importance of tailoring beta scheduling in diffusion models to account for pixel redundancy in the input, which directly correlates with the upsampling scale.
Since each $\eta$ is determined empirically, it provides a robust value for each scale, making it applicable for future work.

\input{partial/sup_fig_beta_search_ablation}




\subsection{Shortcut timestep}

We also conduct an ablation study on shortcut timestep to find the optimal timestep to preserve the image fidelity while enhancing efficiency.
Since our pipeline is based on progressive upsampling, it is important to find the optimal shortcut timestep at the first stage (2048$\times$2048). 
As shown in \Cref{supfig:shortcut}, the quality of generated image restored by diffusion and denoising process using over than 30 timesteps in total 50 timesteps is converged, though the quality of image generated by less than 30 timesteps shows significant degradations.

\input{partial/sup_fig_shortcut}

%% file: partial/sup_tab_statistical_matching.tex
\begin{table}[]
\centering
\vspace{-0.5em}
\renewcommand{\arraystretch}{1.3}
\resizebox{0.85\linewidth}{!}{%

\begin{tabular}{l|l|ll}
\toprule
                             & method & FID$_{256}$$\downarrow$ 
                             & KID $_{256}$$\downarrow$ \\ \hline
\multirow{3}{*}{2048 x 2048} & DemoFusion (30/50) & 44.22 & 0.0198 \\
                             & SM w/ NN            &  \underline{39.64} &  \underline{0.0161} \\
                             & SM w/ Bicubic       & \textbf{39.61} &  \textbf{0.0162}  \\ \midrule
\multirow{3}{*}{3072 x 3072} & DemoFusion (30/50)  &  47.70 &  0.0207 \\
                             & SM w/ NN            & \underline{40.92} &  \textbf{0.0156} \\
                             & SM w/ Bicubic       &  \textbf{39.97} &  \textbf{0.0156}  \\ \midrule
\multirow{3}{*}{4096 x 4096} & DemoFusion (30/50) &  41.50 &  0.0158 \\
                             & SM w/ NN            & \underline{38.34} & \underline{0.0130} \\
                             & SM w/ Bicubic       & \textbf{37.28} & \textbf{0.0127} \\ \bottomrule
\end{tabular}
}%
\caption{
    \textbf{Quantitative results of ablation study on Statistical Matching.} 
    The table presents the results of an ablation study conducted across three resolutions. The highest score for each metric is highlighted in bold, while the second-best score is underlined.
}  
\label{sup_tab:sm_ablation}
\end{table}

%% file: partial/sup_tab_beta_searching.tex
\begin{table}[]
\centering
\small
\vspace{-0.5em}
\renewcommand{\arraystretch}{1.3}
\resizebox{\linewidth}{!}{
\begin{tabular}{c|ll|ll|ll} 
\toprule
\multirow{2}{*}{$\eta$} & \multicolumn{2}{c|}{2048$\times$2048 (scale=2.0)} & \multicolumn{2}{c|}{3072$\times$3072 (scale=1.5)} & \multicolumn{2}{c}{4096$\times$4096 (scale=1.3)} \\  
                        & FID$_{299}$$\downarrow$ & KID$_{299}$$\downarrow$ & FID$_{299}$$\downarrow$ & KID$_{299}$$\downarrow$ & FID$_{299}$$\downarrow$ & KID$_{299}$$\downarrow$ \\  
\midrule
6   & 43.10 & 0.0188 & 44.62 & 0.0187 & 41.05 & 0.0151  \\  
5.5 & 41.20 & 0.0174 & 41.92 & 0.0172 & 38.93 & 0.0140  \\  
5   & 39.33 & 0.0162 & 39.97 & 0.0156 & 36.72 & 0.0125  \\  
4.5 & 38.08 & 0.0151 & 38.27 & 0.0146 & 35.84 & 0.0124  \\  
4   & 36.98 & 0.0140 & 37.30 & 0.0139 & \underline{35.32} & \underline{0.0119}  \\  
3.5 & 36.32 & 0.0136 & 36.50 & 0.0133 & \cellcolor[gray]{0.9}\textbf{35.06} & \cellcolor[gray]{0.9}\textbf{0.0118}  \\  
3   & 35.26 & 0.0129 & \cellcolor[gray]{0.9}\textbf{36.25} & \cellcolor[gray]{0.9}\textbf{0.0131} & 35.73 & 0.0122  \\  
2.5 & 34.88 & 0.0122 & \underline{36.34} & \underline{0.0132} & 37.36 & 0.0136  \\  
2   & \cellcolor[gray]{0.9}\textbf{35.13}  & \cellcolor[gray]{0.9}\textbf{0.0119} & 37.20 & 0.0135 & 38.42 & 0.0138  \\  
1.5 & 35.45 & 0.0121 & 38.42 & 0.0144 & 41.02 & 0.0152  \\  
1   & 35.89 & 0.0120 & 40.28 & 0.0152 & 44.38 & 0.0169  \\  
\bottomrule
\end{tabular}
}%
\caption{
    \textbf{Quantitative results of ablation study on Scale-aware Scheduling.}
    The table presents $\text{FID}_{299}$ and $\text{KID}_{299}$ values for different $\eta$ settings across various upscaling scales. The best score for each resolution is highlighted in bold, and the second-best is underlined.
}
\vspace{-1.5em}
\label{sup_tab:beta_searching}
\end{table}

%% file: partial/sup_fig_snr.tex
\begin{figure}[t]
    \centering
    \includegraphics[width=\linewidth]{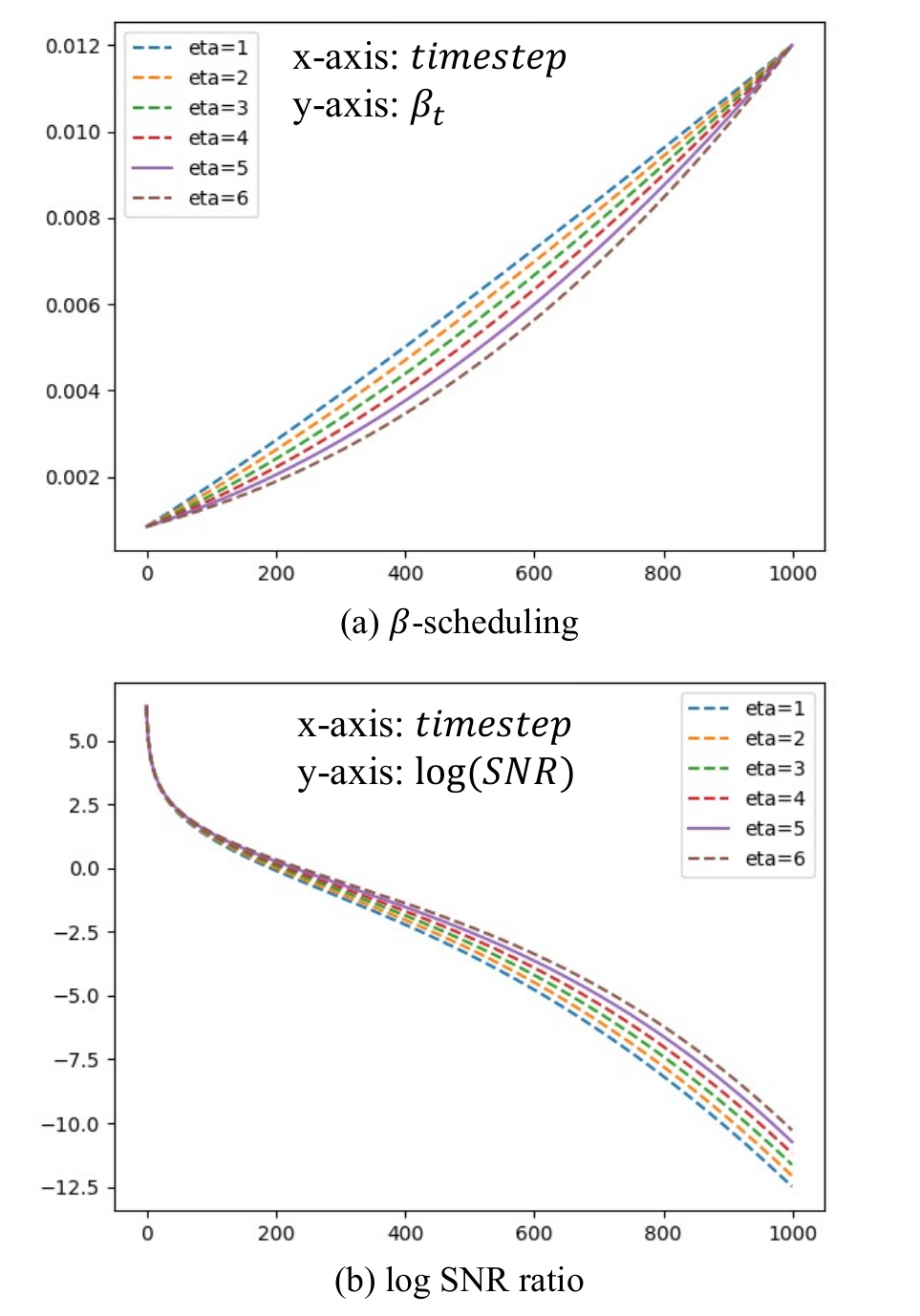}
    \vspace{-1.9em}
    \caption{
        \textbf{Variations of beta scheduling and log signal to noise ratio.}
        The plots show how $\beta_t$ (top) and log(SNR) ratio (bottom) evolve across timesteps under different $\eta$ values, highlighting their impact on noise scheduling and SNR decay during diffusion. The solid line indicates the default beta scheduling of pre-trained diffusion model.
    }
    \label{supfig:snr}
\end{figure}

%% file: partial/sup_fig_beta_search_ablation.tex
\begin{figure*}[htbp]
    \centering
    \includegraphics[width=0.98\textwidth]{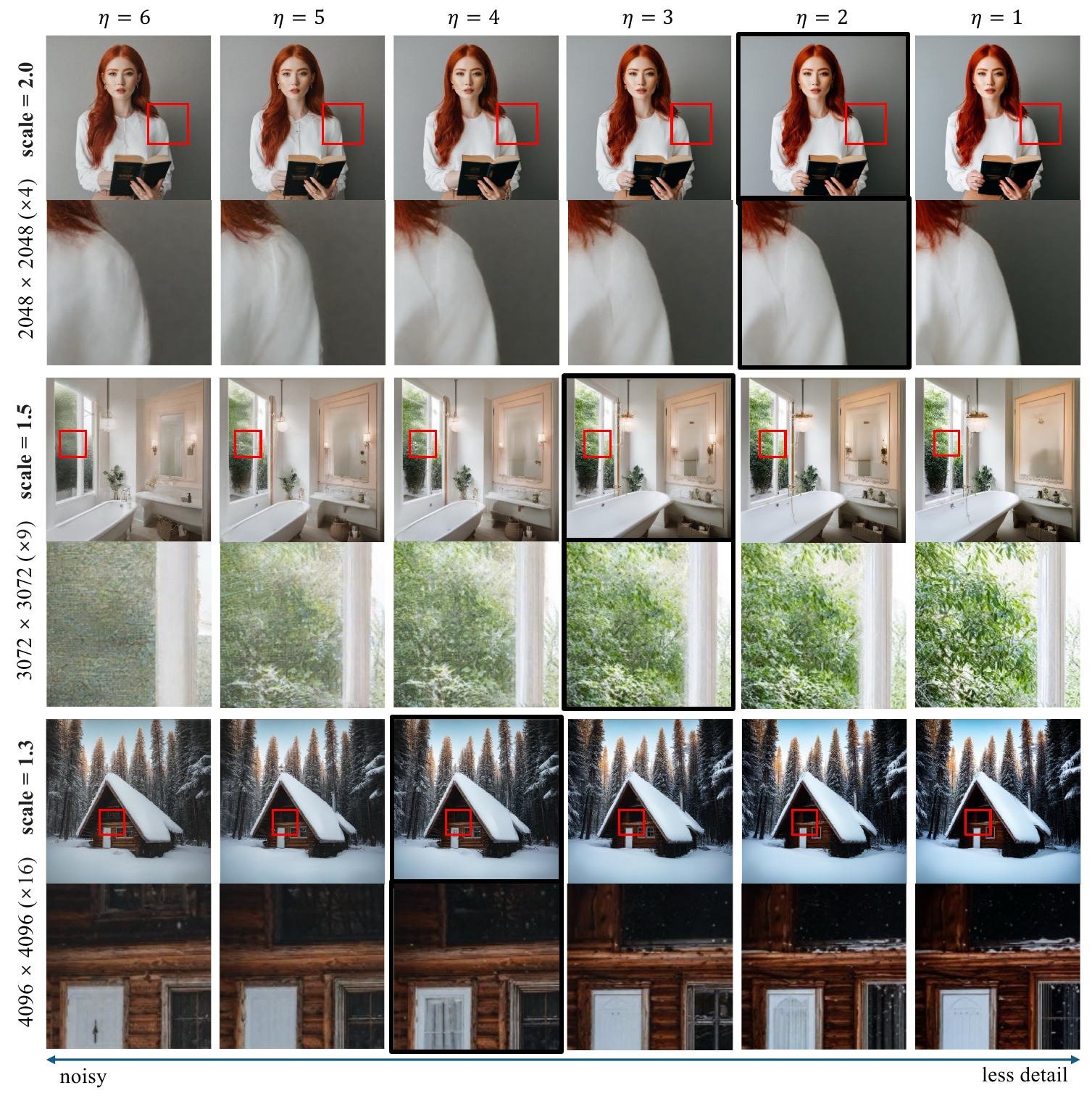}
    \vspace{-0.6em}
    \caption{
        \textbf{Qualitative results of ablation study on Scale-aware Scheduling.}
        The figure compares the visual effects of varying $\eta$ values across three upscaling scales (scale = 2.0, 1.5, 1.3). Each row represents results for a specific scale, showing progression from noisy outputs at higher $\eta$ values to overly smooth outputs with loss of detail at lower $\eta$ values. Red boxes highlight regions for zoomed-in detail examination, illustrating the trade-off between noise reduction and fine detail preservation. Black boxes highlight the results with optimal $\eta$ for each scale.
        }
    \label{sup_fig:beta_search_ablation}
    \vspace{-1em}
\end{figure*}

%% file: partial/sup_fig_shortcut.tex
\begin{figure*}[htbp]
    \centering
    \includegraphics[width=0.98\textwidth]{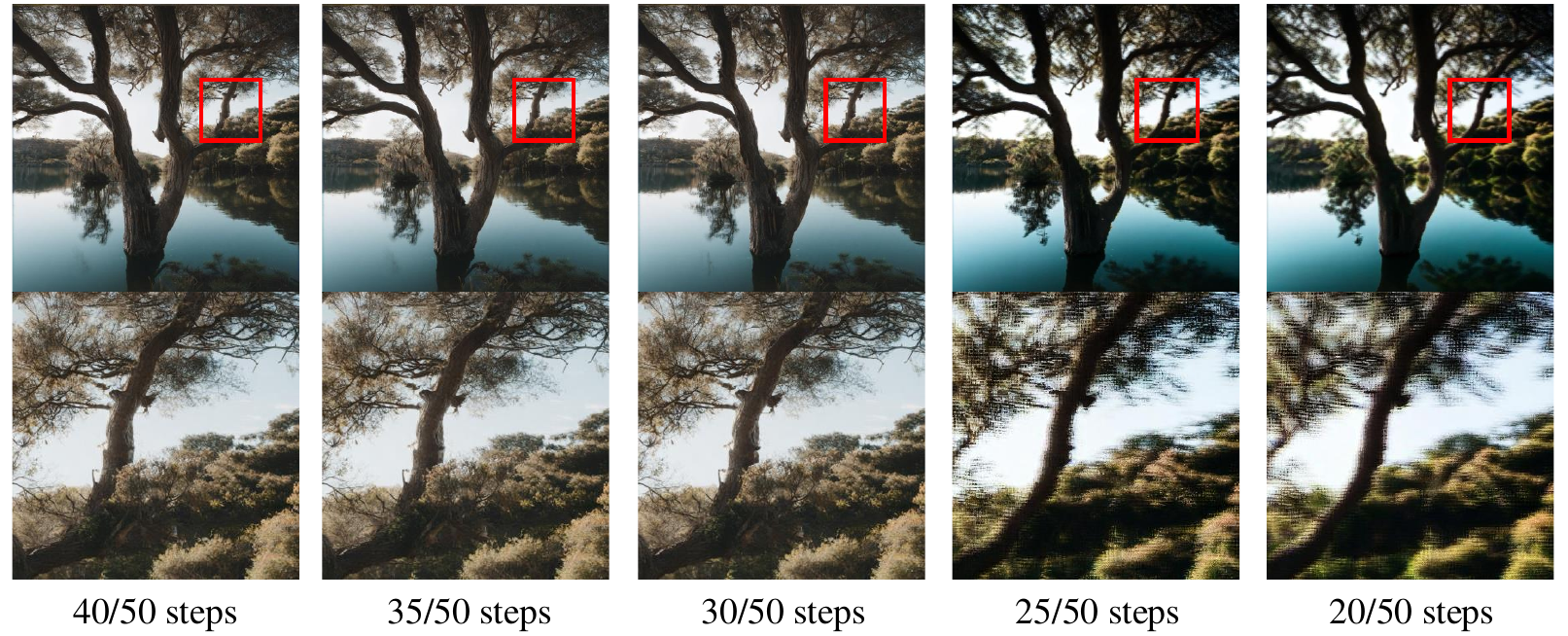}
    \vspace{-0.6em}
    \caption{
        \textbf{Qualitative comparison along to shortcut timestep.}
        This figure illustrates the impact of varying shortcut timesteps (20/50 to 40/50 steps) on the visual quality of generated images. As the number of shortcut timesteps decreases, image quality gradually degrades, with noticeable artifacts and reduced clarity in fine details. The results highlight the trade-off between sampling efficiency and image fidelity.
    }
    \label{supfig:shortcut}
    \vspace{-1.5em}
\end{figure*}

%% file: section/D_Interpolation_upsampling.tex
\section{Detailed motivations of statistical matching}
\label{sup_sec:motivation_of_sm}

Our key observations center on the patch-level distribution shifts. In Section 4.3 in the main paper, we examine how changes in statistical factors (i.e., mean and variance) contribute to image quality degradation. We propose a normalization method, Statistical Matching (SM), to align the mean and variance between the reference latent and the dilated patches from the upsampled latent, effectively handling these distortions.

In this section, we demonstrate the need for SM by introducing how the statistical factors (\ie, mean and variance) of upsampled latent $z^\text{HR}_0\in\mathbb{R}^{\text{H}\times\text{W}\times\text{c}}$ differ from those of the original latent $z_0\in\mathbb{R}^{\text{h}\times\text{w}\times\text{c}}$ where $H>h$ and $W>w$ and how the difference affects to the diffusion process. 
We focus on the dilated patches $d^k_0\in\mathbb{R}^{\text{h}\times\text{w}\times\text{c}}$ from $z^\text{HR}_0$, where $k$ is the index of the patch.

\subsection{Distribution shift in dilated patch}
\label{sup_sec:distribution_shift}

To verify the distribution shifts, we extract the mean and variance of reference latent and those of dilated patches from $\times4$ bicubic upsampled latent. 
To generate reference latents, we utilize various prompts from our testset. 

In \Cref{supfig:stat_var}, the histogram on the top shows that variance of each  dilated patch is reduced compared to that of corresponding reference latent. The reductions are caused by the property of interpolation (\ie, bicubic) which calculate the new pixel value by intermediate value of referring to pixels.
Also, the histogram on the bottom shows that mean of each dilated patch is shifted compared to that of corresponding reference latent. These experimental results demonstrate the necessity of SM onto dilated patches to make them be better initialized as input of the diffusion model.

\input{partial/sup_fig_stat_var}

\subsection{Relation between latent variance and diffusion process}
\label{sup_sec:relation_between_variance_and_diffusion}

Here, we explain the impact of latent variance on the diffusion process.
As described in \Cref{sup_sec:distribution_shift}, the variance of dilated patches from higher resolution latent is reduced due to interpolation-based upsampling. These changes can affect successive deviation from diffusion paths during the generation process.
We demonstrate how the deviation can be occured as follows.

Let $z'_0 = \frac{1}{k} z_0$ be a scaled version of the latent $z_0$ where $k\in\mathbb{R}$, \textit{s.t.} $k\geq1$, is a scaling factor that reduces the variance. This scaled latent $z'_0$ affects the entire diffusion process, resulting in noisy latents $z'_t$ at each timestep $t=1, \dots ,T$ defined by:
\begin{align}
    q(z'_t|z'_{t-1}) &= \mathcal{N}(\sqrt{1-\beta_t}z'_{t-1}, \beta_t\mathbf{I})\\
                 &= \mathcal{N}(\frac{\sqrt{1-\beta_t}}{k}z_{t-1}, \beta_t\mathbf{I}),
\end{align}
where $\beta_t$ is the noise level at each timestep. 
The signal strength $\frac{\sqrt{1-\beta_t}}{k} z_t$ is reduced, disrupting the pre-designed SNR and variance preservation~\cite{ho2020denoising}. 
It also causes the model to diverge from its intended beta schedule, resulting in less stable and lower-quality outputs.

%% file: partial/sup_fig_stat_var.tex
\begin{figure}[t]
    \centering
    \includegraphics[width=0.9\linewidth]{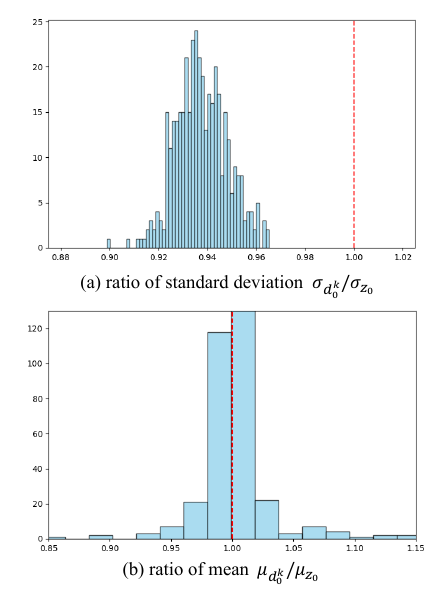}
    \vspace{-0.8em}
    \caption{
        \textbf{Distribution of mean and standard deviation after upsampling.}
        The red line represents the mean and standard deviation of the reference latent, highlighting the deviation in statistical parameters introduced by the upsampling process. The variation of each factor can be found due to upsampling.
        }
    \label{supfig:stat_var}
    \vspace{-1.5em}
\end{figure}

%% file: section/E_SimpleDiffusion.tex
\section{Applying Simple Diffusion to patch-based method}
\label{supp_sec:limitation}

Our approach differs from Simple Diffusion by focusing on pixel redundancy at a fixed resolution and its effect on noise scheduling.
Simple Diffusion suggests that diffusion models should adopt different noise scheduling depending on resolution. 
Given the signal-to-noise ratio ($SNR$) of a latent $z_0\in\mathbb{R}^{s\times s\times c}$, a higher-resolution latent $Z_0\in\mathbb{R}^{S\times S\times c}$ should follow $SNR(t) \times (\frac{s}{S})^2$, where $t$ is the timestep.

There are two ways to apply Simple Diffusion to patch-based high-resolution image generation. One treats each patch as an unique image, making DemoFusion inherently compatible with this method. The other considers a patch as part of a higher resolution image, allowing adjustment of the $(\frac{s}{S})^2$ factor (e.g., $\frac{1}{4}$ for a 2K image).

Figure 4 in the main paper presents the results for both cases. DemoFusion aligns with the first approach, as it preserves the pre-trained beta scheduling for fixed-size local patches. However, this results in distorted fine details (e.g., grass textures). Alternatively, modifying DemoFusion’s beta scheduling according to Simple Diffusion, referred to as DemoFusion+Simple Diffusion, enhances fine details like grass but introduces unnatural artifacts that significantly degrade visual quality.

In contrast, DemoFusion+APT achieves improved results, preserving fine local details without introducing distortions. This highlights the necessity of our method, demonstrating that noise scheduling strategies should account for pixel redundancy in high-resolution image generation.

%% file: section/F_FID.tex
\input{partial/sub_tab_fid_kid}

\section{Analysis of FID for high-resolution image}
\label{supp_sec:weakness_at_fid}

Our method shows marginal performance drop in FID and KID. However, as discussed in the main paper, we argue that these metrics are not proper to evaluate high-resolution image generation models.
Because of InceptionNet, the base embedding model for FID and KID, processes images at a size of 299$\times$299, the high-resolution images evaluated in our experiments are downsampled by up to $13^2$ times, leading to the loss of majority high frequency details.

To further support our claim, we refer to \Cref{sup_tab:fid_kid}. Notably, DemoFusion (30/50) achieves better FID and KID scores than DemoFusion (50/50), despite of its blurrier and less detailed images. This pattern aligns with observations from results of super resolution using naive interpolation based methods, where the method achieving the best FID and KID scores.
Furthermore, given the clear qualitative differences in detail quality as shown in Figure 1. and Figure 6. in the main paper, we argue that FID and KID should not be treated as primary metrics for our task. 
This trend has also been reported in several previous works~\cite{podellsdxl, kirstain2023pick}.

%% file: partial/sub_tab_fid_kid.tex
\begin{table}[]
\centering
\begin{tabular}{l|cc}
\hline
                   & \multicolumn{1}{l}{FID} & \multicolumn{1}{l}{KID} \\ \hline
SDXL+Bicubic       & 82.92                   & 0.0065                  \\
SDXL+LANCZOS       & 82.92                   & 0.0065                  \\
DemoFusion         & 86.24                   & 0.0083                  \\
DemoFusion (30/50) & 85.03                   & 0.0080                  \\
DemoFusion+APT     & 87.18                   & 0.0091                  \\ \hline
\end{tabular}
\caption{
    \textbf{Analysis of FID and KID for high resolution image generation.} 
}
\label{sup_tab:fid_kid}
\end{table}

%% file: section/G_Extreme_Resolution.tex
\section{Additional qualitative results}
\label{supp_sec:limitation}

We show additional qualitative results in \Cref{supfig:accdiffusion} and \Cref{rebut:8k}.

\input{partial/sup_fig_accdiffusion}
\input{partial/rebut_fig_8k}

\clearpage

%% file: partial/sup_fig_accdiffusion.tex
\begin{figure*}[]
    \centering
    \includegraphics[width=\linewidth]{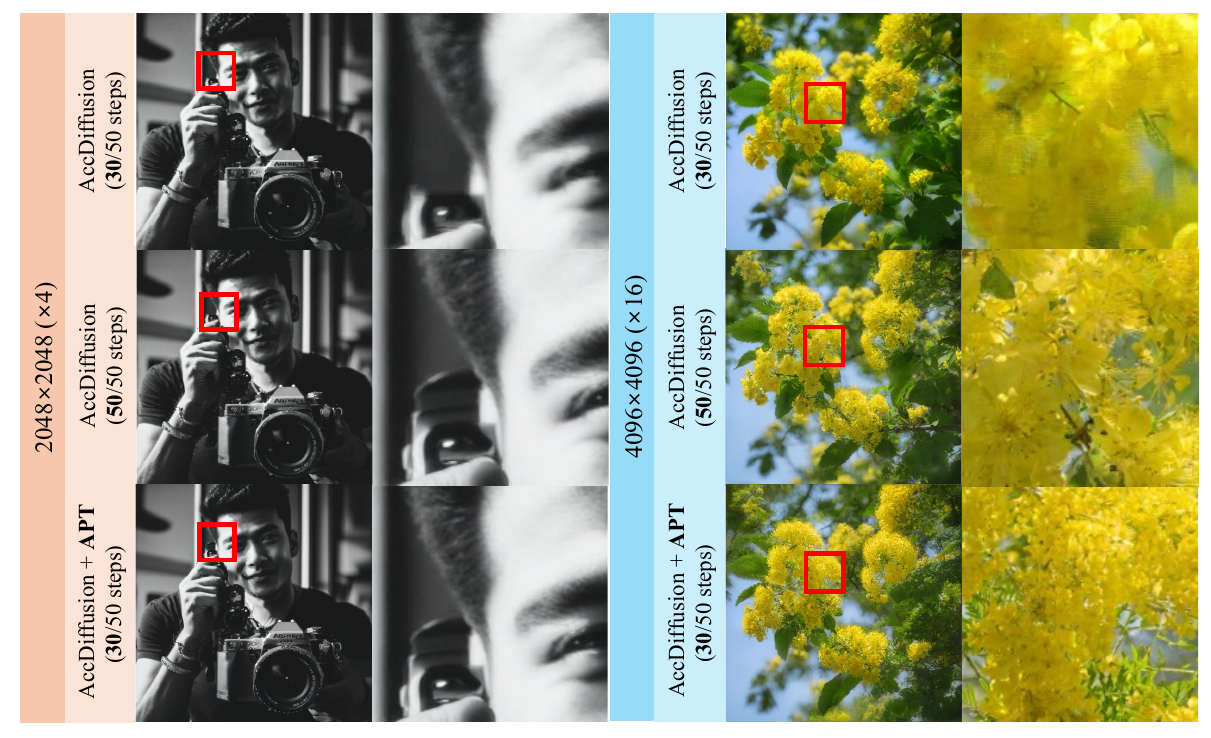}
    \vspace{-1.9em}
    \caption{
        \textbf{Qualitative results of APT applied to AccDiffusion.} 
        This figure compares the visual quality of high-resolution images generated by AccDiffusion (30/50 steps) and AccDiffusion combined with APT. At both resolutions, APT enhances fine details and reduces visual artifacts, as highlighted in the zoomed-in regions.
    }
    \label{supfig:accdiffusion}
\end{figure*}



%% file: partial/rebut_fig_8k.tex
\begin{figure*}[t]
    \centering
    \includegraphics[width=\linewidth]{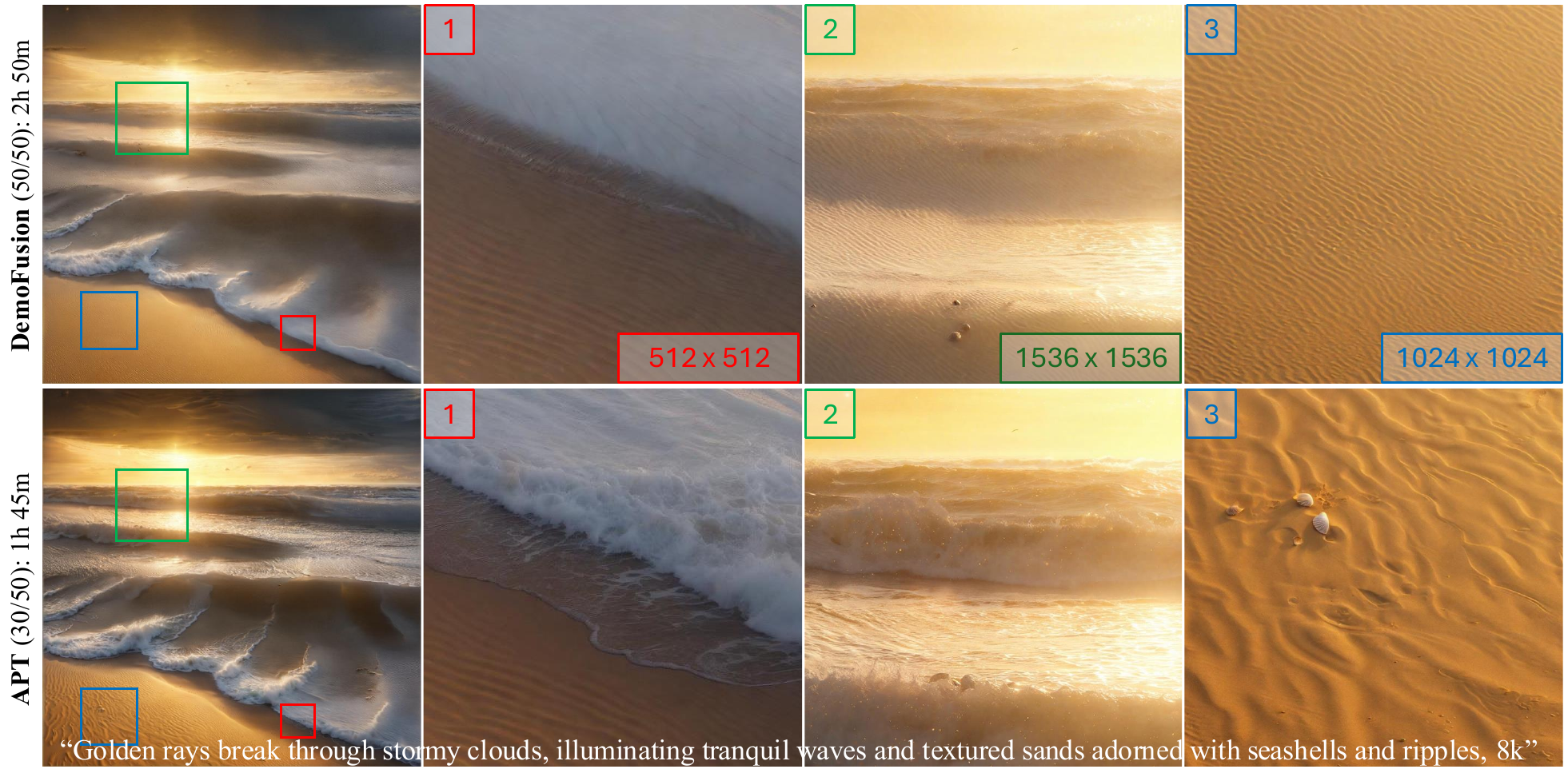}
    \caption{
       \textbf{Examples for extreme resolution.} Generated images at resolution 8K by DemoFusion and APT with a prompt “Golden rays break through stormy clouds, illuminating tranquil waves and textured sands adorned with seashells and ripples, 8k”.
    }
    \label{rebut:8k}
\end{figure*}


